%% file: zmain.tex
\documentclass[10pt,twocolumn,letterpaper]{article}

\usepackage{packages/wacv}
\usepackage{times}
\usepackage{epsfig}
\usepackage{graphicx}
\usepackage{amsmath}
\usepackage{amssymb}
\usepackage{booktabs}
\usepackage{mathtools}
\usepackage{subcaption}
\usepackage{xcolor}
\usepackage{multirow}

\usepackage[accsupp]{axessibility}  
\DeclarePairedDelimiter{\floor}{\lfloor}{\rfloor}

\definecolor{mygreen}{RGB}{60, 186, 84}
\definecolor{myred}{RGB}{219, 50, 54}
\definecolor{myblack}{RGB}{0,0,0}

\def\wacvPaperID{228} 

\wacvfinalcopy 

\ifwacvfinal
\usepackage[breaklinks=true,bookmarks=false]{hyperref}
\else
\usepackage[pagebackref=true,breaklinks=true,colorlinks,bookmarks=false]{hyperref}
\fi

\usepackage{multibib}
\newcites{Supp}{References}

\begin{document}

\title{Contrastive Losses Are Natural Criteria for Unsupervised Video Summarization}

\author{Zongshang Pang \\
Osaka University \\
{\tt\small pangzs@is.ids.osaka-u.ac.jp} 
\and
Yuta Nakashima \\
Osaka University \\
{\tt\small n-yuta@ids.osaka-u.ac.jp}
\and
Mayu Otani \\
CyberAgent, Inc. \\
{\tt\small otani-mayu@cyberagent.co.jp}
\and
Hajime Nagahara \\
Osaka University \\
{\tt\small nagahara@ids.osaka-u.ac.jp}
}

\maketitle
\thispagestyle{empty}
\input{Abstract}

\input{introduction.tex}

\input{relatedwork.tex}

\input{preliminaries.tex}

\input{method.tex}

\input{experiment.tex}

\input{conclusion.tex}

\input{acknowledgement.tex}
{\small
\bibliographystyle{packages/ieee_fullname}
\bibliography{packages/egbib}
}

\input{supp.tex}
\clearpage
{\small
\bibliographystyleSupp{packages/ieee_fullname}
\bibliographySupp{packages/egbib_supp}
}

\end{document}

%% file: abstract.tex
\begin{abstract}
Video summarization aims to select the most informative subset of frames in a video to facilitate efficient video browsing. Unsupervised methods usually rely on heuristic training objectives such as diversity and representativeness. However, such methods need to bootstrap the online-generated summaries to compute the objectives for importance score regression. We consider such a pipeline inefficient and seek to directly quantify the frame-level importance with the help of contrastive losses in the representation learning literature. Leveraging the contrastive losses, we propose three metrics featuring a desirable key frame: local dissimilarity, global consistency, and uniqueness. With features pre-trained on the image classification task, the metrics can already yield high-quality importance scores, demonstrating competitive or better performance than past heavily-trained methods. We show that by refining the pre-trained features with a lightweight contrastively learned projection module, the frame-level importance scores can be further improved, and the model can also leverage a large number of random videos and generalize to test videos with decent performance. Code available at \url{https://github.com/pangzss/pytorch-CTVSUM}.
\end{abstract}

%% file: introduction.tex
\section{Introduction}

Recently, deep neural networks have significantly advanced the development of efficient video summarization tools. The supervised workflow and evaluation protocols proposed by Zhang \etal~\cite{zhang2016video} have become a cornerstone for most of the subsequent deep-learning-based supervised methods. Unsupervised methods avoid using annotated summaries by utilizing heuristic training objectives such as \textit{diversity} and \textit{representativeness} \cite{mahasseni2017unsupervised,rochan2018video,liu2019learning,rochan2019video,zhou2018deep,jung2019discriminative,jung2020global}. The diversity objective aims to enforce the dissimilarity between the key frame candidates, and the representativeness objective guarantees that the generated summaries can well reflect the major information in the original video.



The past unsupervised approaches focus on bootstrapping the summaries generated on the fly during training to evaluate their diversity and representativeness and then utilizing the resulting loss terms to train models. However, the basis for these algorithms is, by all means, the more fundamental elements of the summaries, \ie, the frames. The premise for producing a decent summary is that the correct frames are selected. Bootstrapping the online-generated summaries with poor quality seems less straightforward, if not redundant. Here we pose a question: \textit{How do we directly quantify how much each frame contributes to the quality of the final summary?} 

\begin{figure}
\includegraphics[width=1 \linewidth]{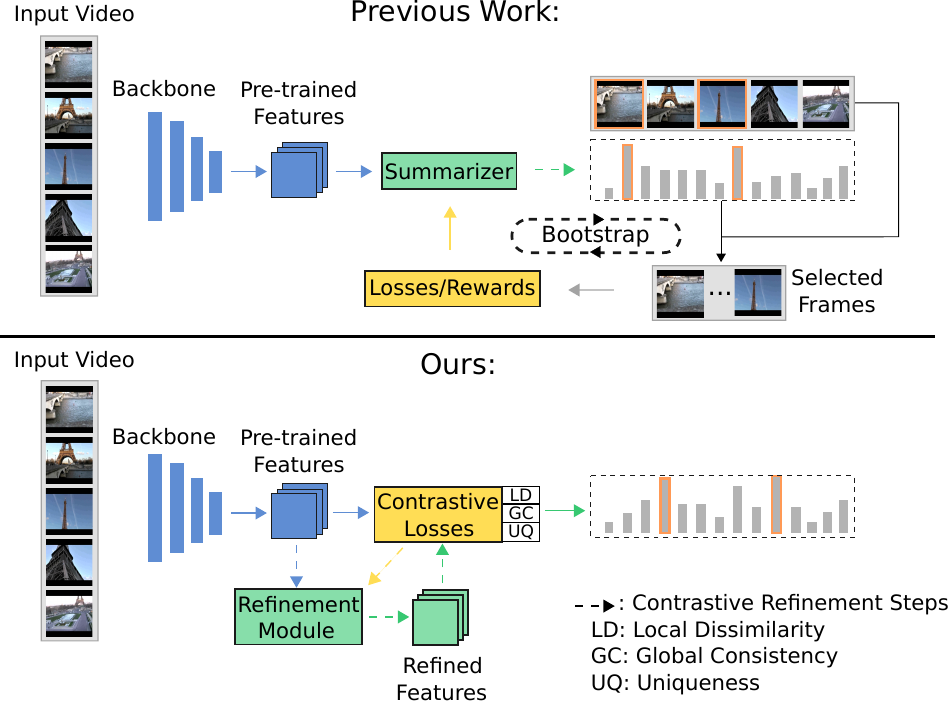}
\caption{A comparison between our method and previous work.} 
\label{fig1}
\end{figure}

To answer this question, we start by defining two desirable properties for key frame candidates: \emph{local dissimilarity} and \emph{global consistency}. Inspired by the diversity objective, if a frame is excessively similar to its semantically close neighbors in the feature space, then this frame, together with its neighbors, lacks \textit{local} dissimilarity, where the locality is defined in the feature space based on cosine similarity \cite{zhuang2019local}. The information in such frames is often monotonous, as they appear multiple times in the video but rarely demonstrate any variations. Thus, we risk introducing redundancy to the final summary if such frames are considered key frames. On the other hand, merely selecting frames based on dissimilarity may wrongly incorporate noisy frames with very few semantically meaningful neighbors, thus not indicative of the video theme. Inspired by the representativeness objective, we consider frames consistent with the majority of frames in a video as being related to the central video theme, \ie, they are globally consistent. Eventually, we would like to select frames with a desirable level of local dissimilarity and global consistency so that the resulting summary can have well-balanced diversity and representativeness.

Interestingly, the above two metrics can be readily calculated by utilizing contrastive losses for image representation learning, \ie, alignment and uniformity \cite{wang2020understanding} losses. The alignment loss calculates the distance between an image and a semantically related sample, \eg, an augmented version of the image. With a pool of semantically related samples, Zhuang \etal~\cite{zhuang2019local} defines the aggregation of these samples as a local neighborhood. The alignment loss can readily measure the local dissimilarity of each frame in such a neighborhood. The uniformity loss aims to regularize the proximity of the overall features and would be high for closely distributed features. Therefore, it can be conveniently leveraged to measure the semantic consistency between frames. The two losses can be further utilized to perform \emph{contrastive refinement} of the features, which we will show is also more efficient than previous methods. 

Nonetheless, background frames with complex contents related to many other frames in the video can also be locally dissimilar and globally consistent. For instance, street scenes will likely appear throughout a car-accident video. Such frames can still be relatively dissimilar because of the moving objects. However, on average, they may be consistent with most frames. Luckily, we can exclude them by leveraging an assumption that these background frames tend to appear in many different videos and thus are not \textit{unique} to their associated videos, \eg,  street scenes in videos about car accidents, parades, city tours, \etc. Based on this assumption, we train a \emph{unqueness filter} to filter out ambiguous background frames, which can be readily incorporated into the aforementioned contrastive losses. We illustrate our proposed method together with a comparison with previous work in Fig.~\ref{fig1}. 

\textbf{Contributions.} Unlike previous work bootstrapping the online-generated summaries, we propose three metrics, local dissimilarity, global consistency, and uniqueness to directly quantify frame-level importance based on theoretically motivated contrastive losses \cite{wang2020understanding}. This is significantly more efficient than previous methods. Specifically, we can obtain competitive F1 scores and better correlation coefficients \cite{otani2019rethinking} on SumMe \cite{gygli2014creating} as well as TVSum \cite{song2015tvsum} with the first two metrics calculated using only ImageNet \cite{krizhevsky2012imagenet} pre-trained features \textit{without any further training}. Moreover, by contrastively refining the features along with training the proposed uniqueness filter, we can further improve the performance given only random videos sampled from the Youtube8M dataset \cite{abu2016youtube}

%% file: relatedwork.tex
\section{Related Work}
Before the dawn of deep learning, video summarization was handled by hand-crafted features and heavy optimization schemes \cite{song2015tvsum,gygli2015video,liu2010hierarchical,zhao2014quasi,gygli2014creating}. Zhang \etal~\cite{zhang2016video} applied a deep architecture involving a bidirectional recurrent neural network (RNN) to select key frames in a supervised manner, forming a cornerstone for many subsequent work \cite{zhao2017hierarchical,zhao2018hsa,zhang2018retrospective,feng2018extractive,wang2019stacked}. There are also methods that leverage attention mechanisms \cite{fajtl2018summarizing,casas2019video,ji2019video,ji2020deep,liu2019learning,fu2019attentive,liu2020transforming,lin2020bi} or fully convolutional networks \cite{rochan2018video, long2015fully}. Exploring spatiotemporal information by jointly using RNNs and convolutional neural networks (CNNs) \cite{yuan2019spatiotemporal,chu2019spatiotemporal,elfeki2019video} or using graph convolution networks \cite{kipf2016semi,park2020sumgraph} also delivered decent performance. There is also an attempt at leveraging texts to aid video summarization \cite{narasimhan2021clip}.

Unsupervised methods mainly exploit two heuristics: diversity and representativeness. Zhou \etal~\cite{zhou2018deep} and subsequent work \cite{chen2019weakly,li2021weakly} tried to maximize the diversity and representativeness rewards of the produced summaries. Some work \cite{mahasseni2017unsupervised, park2020sumgraph,rochan2018video,rochan2019video} applied the repelling regularizer \cite{zhao2016energy} to regularize the similarities between the mid-level frame features in the generated summaries to guarantee their diversity. Mahasseni \etal~\cite{mahasseni2017unsupervised} and subsequent work \cite{jung2019discriminative,he2019unsupervised,liu2019learning} used the reconstruction-based VAE-GAN structure to generate representative summaries, while Rochan \etal~\cite{rochan2019video} exploited reconstructing unpaired summaries.

Though also inspired by diversity and representativeness, our methodology differs from all the above unsupervised approaches. Concretely, we directly quantify how much each frame contributes to the final summary by leveraging contrastive loss functions in the representation learning
literature \cite{wang2020understanding}, which enables us to achieve competitive or better performance compared to these approaches \emph{without any training}. Moreover, we perform contrastive refinement of the features to produce better importance scores, which obviates bootstrapping online generated summaries and is much more efficient than previous work. To the best of our knowledge, we are the first to leverage contrastive learning for video summarization.

%% file: preliminaries.tex
\section{Preliminaries}
As contrastive learning is essential to our approach, we introduce the preliminaries focusing on instance discrimination \cite{wu2018unsupervised}.

\subsection{Instance Discrimination via the InfoNCE Loss}

As an integral part of unsupervised image representation learning, contrastive learning \cite{chopra2005learning} has been attracting researchers' attention over the years and has been constantly improved to deliver representations with outstanding transferability \cite{wu2018unsupervised, he2019unsupervised, zhuang2019local, wang2020understanding, hjelm2018learning, oord2018representation, tian2020contrastive, chen2020simple}.
Formally, given a set $\mathcal{D}=\{I_n\}_{n=1}^N$ of $N$ images, contrastive representation learning aims to learn an encoder $f_{\theta}$ with learnable $\theta$ such that the resulting features $f_{\theta}(I_n)$ can be readily leveraged by downstream vision tasks. A theoretically founded \cite{oord2018representation} loss function with favorable empirical behaviors \cite{wang2021understanding} is the so-called InfoNCE loss \cite{oord2018representation}:
\begin{equation}
    \mathcal{L}_{\text{InfoNCE}} = \sum_{I \in \mathcal{D}} -\log
    \frac{e^{f_{\theta}(I) \cdot f_{\theta}(I')/\tau}}{\sum_{J \in \mathcal{D}'(I)}e^{f_{\theta}(I) \cdot f_{\theta}(J) / \tau }}, \label{eq1}
\end{equation}
where $I'$ is a positive sample for $I \in X$, usually obtained through data augmentation, and $\mathcal{D}'(I)$ includes $I'$ as well as all negative samples, \eg, any other images. The operator ``$\cdot$'' is the inner product, and $\tau$ is a temperature parameter. Therefore, the loss aims at pulling closer the feature of an instance with that of its augmented views while repelling it from those of other instances, thus performing instance discrimination.

\subsection{Contrastive Learning via Alignment and Uniformity}\label{sec3.2}
When normalized onto the unit hypersphere, the contrastively learned features that tend to deliver promising downstream performance possess two interesting properties. That is, the semantically related features are usually closely located on the sphere regardless of their respective details, and the overall features' information is retained as much as possible \cite{oord2018representation, hjelm2018learning, tian2020contrastive}. Wang \etal~\cite{wang2020understanding} defined these two properties as \textit{alignment} and \textit{uniformity}.

The alignment metric computes the distance between the positive pairs \cite{wang2020understanding}:
\begin{equation}\label{eq2}
    \mathcal{L}_{\text{align}}(\theta, \alpha) = \underset{{(I,I')\sim p_{\text{pos}}}}{\mathbb{E}}[\| f_{\theta}(I) - f_{\theta}(I') \|^{\alpha}_{2}], 
\end{equation}
where $\alpha > 0$, and $p_\text{pos}$ is the distribution of positive pairs (\ie, an original image and its augmentation).

The uniformity is defined as the average pairwise Gaussian potential between the overall features:
\begin{equation}\label{eq3}
    \mathcal{L}_{\text{uniform}}(\theta, \beta) = \log \left(\underset{I,J\overset{\text{i.i.d}}{\sim} p_{\text{data}}}{\mathbb{E}}[e^{-\beta \|f_{\theta}(I) - f_{\theta}(J)\|^{2}_{2}}]\right), 
\end{equation}
where $p_{\text{data}}$ is usually approximated by the empirical data distribution, and $\beta$ is usually set to 2 as recommended by \cite{wang2020understanding}.
This metric encourages the overall feature distribution on the unit hypersphere to approach a uniform distribution and can also be directly used to measure the uniformity of feature distributions \cite{wang2021understanding}. Moreover, Eq.~(\ref{eq3}) approximates the log of the denominator of Eq.~(\ref{eq1}) when the number of negative samples goes to infinity \cite{wang2020understanding}. As proved in \cite{wang2020understanding}, jointly minimizing Eqs.~(\ref{eq2}) and (\ref{eq3}) can achieve better alignment and uniformity of the features, \ie, they are locally clustered and globally uniform \cite{wang2021understanding}.

In this paper, we use Eq.~(\ref{eq2}) to compute the distance/dissimilarity between the semantically close video frame features to measure frame importance in terms of local dissimilarity. We then use the proposed variant of Eq.~(\ref{eq3}) to measure the proximity between a specific frame and the overall information of the associated video to estimate their semantic consistency. Moreover, utilizing the two losses, we learn a nonlinear projection of the pre-trained features so that the projected features are more locally aligned and globally uniform.


%% file: method.tex
\begin{figure}
\includegraphics[width=1 \linewidth]{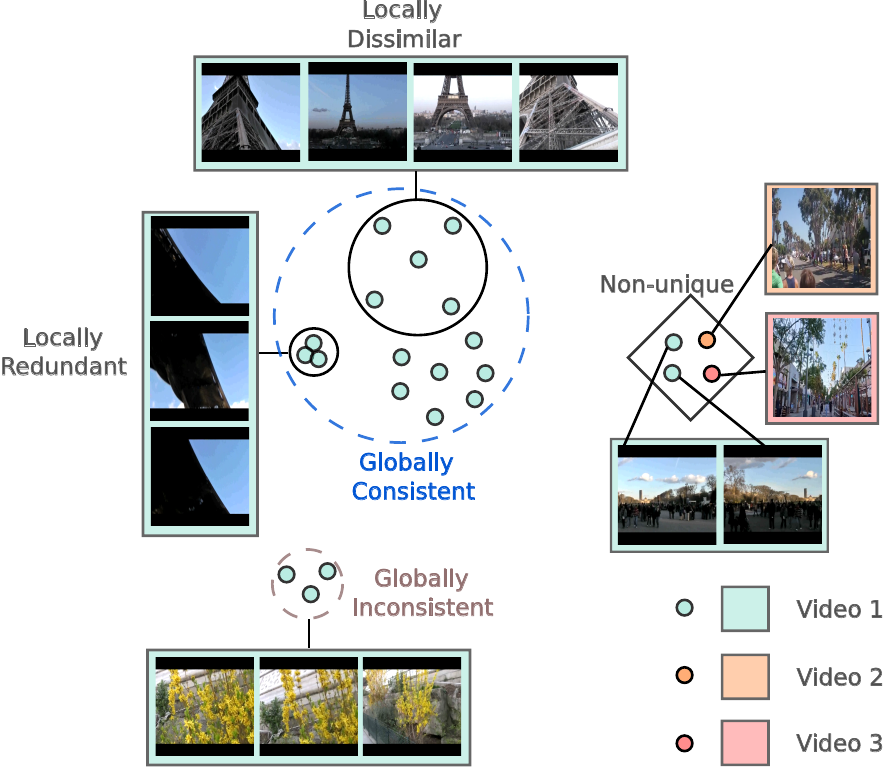}
\caption{A conceptual illustration for the three metrics: local dissimilarity, global consistency, and uniqueness in the semantic space. The images come from the SumMe \cite{gygli2014creating} and TVSum \cite{song2015tvsum} datasets. The dots with the same color indicate features from the same video. For concise demonstration, we only show one frame for ``video 2'' and ``video 3'' to show the idea of uniqueness.}
\label{fig2}
\end{figure}

\section{Proposed Method}\label{section:3}
Unlike previous work that bootstraps the inaccurate candidate summaries generated on the fly, we take a more straightforward perspective to quantify frame importance directly, as we believe it is highly inefficient to deal with the infinitely many collections of frames. To quantify frame importance, we define three metrics: local dissimilarity, global consistency, and uniqueness. We provide a conceptual illustration in Fig.~\ref{fig2}.





\subsection{Local Dissimilarity}

Inspired by the diversity objective, we consider frames likely to result in a diverse summary as those conveying diverse information even when compared to their semantic nearest neighbors. 

Formally, given a video $\mathbf{V}$, we first extract deep features using ImageNet \cite{krizhevsky2012imagenet} pre-trained backbone, \eg, GoogleNet \cite{szegedy2015going}, denoted as $F$, such that $F(\mathbf{V})=\{\mathbf{x}_{t}\}^{T}_{t=1}$, where $\mathbf{x}_{t}$ represents the deep feature for the $t$-th frame in $\mathbf{V}$, and $T$ is the total number of frames in $\mathbf{V}$. Each feature is L2-normalized such that $\|\mathbf{x}_{t}\|_{2}=1$.

To define local dissimilarity for frames in $\mathbf{V}$, we first use cosine similarity to retrieve for each frame $\mathbf{x}_{t}$ a set $\mathcal{N}_{t}$ of top $K = a T$ neighbors, 
where $a$ is a hyperparameter and $K$ is rounded to the nearest integer. The local dissmilarity metric for $\mathbf{x}_{t}$ is an empirical approximation of Eq.~(\ref{eq2}), defined as the local alignment loss:
\begin{equation}\label{eq4}
    \mathcal{L}_{\text{align}}(\mathbf{x}_t) = \frac{1}{|\mathcal{N}_{t}|}\sum_{\mathbf{x} \in \mathcal{N}_{t}}\| \mathbf{x}_{t} - \mathbf{x} \|^{2}_{2},
\end{equation}
which measures the distance/dissimilarity between $\mathbf{x}_{t}$ and its semantic neighbors. 

The larger $\mathcal{L}_{\text{align}}(\mathbf{x}_t)$ is, the more dissimilar $\mathbf{x}_{t}$ is with its neighbors. Thus, if a frame has a certain distance from even its nearest neighbors in the semantic space, the frames in their local neighborhood are likely to convey diverse but still semantically cohesive information and hence are desirable key frame candidates. 
$\mathcal{L}_{\text{align}}(\mathbf{x}_t)$ can be directly used as the importance score of $\mathbf{x}_{t}$ after proper scaling.

\subsection{Global Consistency}
$\mathcal{N}_{t}$ may contain semantically irrelevant frames if $\mathbf{x}_{t}$ has very few semantic neighbors in the video. Therefore, merely using Eq.~(\ref{eq4}) as framewise importance scores is insufficient. 

Inspired by the reconstruction-based representativeness objective \cite{mahasseni2017unsupervised}, we define another metric called global consistency to quantify how consistent a frame is with the video gist by a modified uniformity loss based on Eq.~(\ref{eq3}):
\begin{equation}\label{eq5}
    \mathcal{L}_{\text{uniform}}(\mathbf{x}_t) = \log \left(\frac{1}{T-1}\sum_{\substack{\mathbf{x} \neq \mathbf{x}_t, \\ \mathbf{x} \in F(\mathbf{V})}}e^{-2\| \mathbf{x}_{t} - \mathbf{x} \|^{2}_{2}}\right),
\end{equation}

$\mathcal{L}_{\text{uniform}}(\mathbf{x}_t)$ measures the proximity between $\mathbf{x}_{t}$ and the remaining frames, bearing similarity to the reconstruction- and K-medoids-based objectives in \cite{mahasseni2017unsupervised, zhou2018deep}, but only using a single frame instead of a collection of frames for reconstruction and obviating training an autoencoder \cite{mahasseni2017unsupervised} or a policy network \cite{zhou2018deep}. 

\subsection{Contrastive Refinement}\label{sec4.3}
Eqs. (\ref{eq4}) and (\ref{eq5}) are calculated using deep features pre-trained on image classification tasks, which may not necessarily well possess the local alignment and global uniformity as discussed in Section \ref{sec3.2}. 

Hamilton \etal \cite{hamilton2022unsupervised} contrastively refines the self-supervised vision transformer features \cite{caron2021emerging} for unsupervised semantic segmentation. They freeze the feature extractor for better efficiency and only train a lightweight projector. Inspired by this work, we also avoid fine-tuning the heavy feature extractor, a deep CNN in our case, but instead only train a lightweight module appended to it.

Formally, given features $F(\mathbf{V})$ from the frozen backbone for a video, we feed them to a learnable module to obtain $\mathbf{z}_{t} = G_{\theta}(\mathbf{x}_{t})$, where $\mathbf{z}_{t}$ is L2-normalized\footnote{We leave out the L2-normalization operator for notation simplicity.}. The nearest neighbors in $\mathcal{N}_{t}$ for each frame are still determined using the pre-trained features $\{\mathbf{x}_{t}\}^{T}_{t=1}$, which have been shown to be a good proxy for semantic similarities \cite{ufer2017deep, hamilton2022unsupervised}. Similar to \cite{wang2021dense, zhuang2019local}, we also observed collapsed training when directly using the learnable features for nearest neighbor retrieval, so we stick to using the frozen features.

With the learnable features, the alignment (local dissimilarity) and uniformity (global consistency) losses become \footnote{We slightly abuse the notation of $\mathcal{L}$ to represent losses both before and after transformation by $G_\theta$.}
\begin{align}
    \mathcal{L}_{\text{align}}(\mathbf{z}_{t};\theta) &= \frac{1}{|\mathcal{N}_{t}|}\sum_{\mathbf{z} \in \mathcal{N}_{t}}\| \mathbf{z}_{t} - \mathbf{z} \|^{2}_{2}, \label{eq6} \\
    \mathcal{L}_{\text{uniform}}(\mathbf{z}_{t}; \theta) &= \log \left(\frac{1}{T - 1}\sum_{\substack{\mathbf{z} \neq \mathbf{z}_t, \\ \mathbf{z} \in G_{\theta}(F(\mathbf{V}))}}e^{-2 \| \mathbf{z}_{t} - \mathbf{z} \|^{2}_{2}}\right),\label{eq7}
\end{align}

The joint loss function is thus:
\begin{equation}\label{eq8}
    \mathcal{L}(\mathbf{z}_{t}; \theta) = \mathcal{L}_{\text{align}}(\mathbf{z}_{t}; \theta) + \lambda_1 \mathcal{L}_{\text{uniform}}(\mathbf{z}_{t}; \theta),
\end{equation}
where $\lambda_1$ is a hyperparameter balancing the two loss terms.

During the contrastive refinement, frames that have semantically meaningful nearest neighbors and are consistent with the video gist will have an $\mathcal{L}_{\text{align}}$ and an $\mathcal{L}_{\text{uniform}}$ that mutually resist each other. Specifically, when a nontrivial number of frames beyond $\mathcal{N}_{t}$ also share similar semantic structures with the anchor $\mathbf{z}_{t}$, these frames function as ``hard negatives'' that prevent $\mathcal{L}_{\text{align}}$ to be easily minimized \cite{zhuang2019local, wang2021understanding}. Therefore, only frames with both moderate local dissimilarity and global consistency will have balanced values for the two losses. In contrast, the other frames tend to have extreme values compared to those before the refinement. 



\subsection{The Uniqueness Filter}
The two metrics defined above overlook the fact that the locally dissimilar and globally consistent frames can be background frames with complex content that may be related to most of the frames in the video. For instance, dynamic city views can be ubiquitous in videos recorded in a city. 

Intriguingly, we can filter out such frames by directly leveraging a common property of theirs: They tend to appear in many different videos that may not necessarily share a common theme and may or may not have a similar context, \eg,  city views in videos about car accidents, city tours, and city parades, \etc, or scenes with people moving around that can appear in many videos with very different contexts. Therefore, such frames are not \textit{unique} to their associated videos. Similar reasoning is exploited in weakly-supervised action localization literature \cite{nguyen2019weakly, liu2019completeness, lee2020background}, where a single class is used to capture all the background frames. However, we aim to pinpoint background frames in an unsupervised manner. Moreover, we do not use a single prototype to detect all the backgrounds as it is too restricted \cite{lee2021weakly}. Instead, we treat each frame as a potential background prototype to look for highly-activated frames in random videos, which also determines the backgourndness of the frame itself.

To design a filter for eliminating such frames, we introduce an extra loss to Eq.~(\ref{eq8}) that taps into cross-video samples. For computational efficiency, we aggregate the frame features in a video $\mathbf{V}_k$ with $T_k$ frames into segments with an equal length of $m$. The learnable features $\mathbf{z}$ in each segment are average-pooled and L2-normalized to obtain segment features $\mathcal{S}_k = \{\mathbf{s}_{l}\}_{l=1}^{|\mathcal{S}_k|}$ with $|\mathcal{S}_k|=\floor{T_k/m}$. To measure the proximity of a frame with frames from a randomly sampled batch of videos $\mathcal{B}$ (represented now as segment features) including $\mathcal{S}_k$, we again leverage Eq.~(\ref{eq3}) to define the uniqueness loss for $\mathbf{z}_{t} \in \mathbf{V}_k$ as:
\begin{equation}\label{eq10}
    \mathcal{L}_{\text{unique}}(\mathbf{z}_{t}; \theta) = \log \left(\frac{1}{A} \sum_{\mathcal{S} \in \mathcal{B}/\mathcal{S}_k}\sum_{\mathbf{s} \in \mathcal{S}}e^{-2\| \mathbf{z}_{t} - \mathbf{s}\|^{2}_{2}}\right),
\end{equation}
where $A=\sum_{\mathcal{S} \in \mathcal{B}/\mathcal{S}_k} |\mathcal{S}|$ is the normalization factor. A large value of  $\mathcal{L}_{\text{unique}}$ means that $\mathbf{z}_{t}$ has nontrivial similarity with segments from randomly gathered videos, indicating that it is likely to be a background frame.

When jointly optimized with Eq.~(\ref{eq8}), Eq.~(\ref{eq10}) will be easy to minimize for unique frames, for which most of $\mathbf{s}$ are semantically irrelevant and can be safely repelled. It is not the case for the background frames that have semantically similar $\mathbf{s}$, as the local alignment loss keeps strengthening the closeness of semantically similar features. 

As computing Eq.~(\ref{eq10}) needs random videos, it is not as straightforward to convert Eq.~(\ref{eq10}) to importance scores after training. To address this, we simply train a model $H_{\hat{\theta}}$ whose last layer is sigmoid to mimic $1 - \bar{\mathcal{L}}_{\text{unique}}(\mathbf{z}_t; \theta)$, where $\bar{\mathcal{L}}_{\text{unique}}(\mathbf{z}_t; \theta)$ is $\mathcal{L}_{\text{unique}}(\mathbf{z}_t; \theta)$ scaled to $[0, 1]$ over $t$. Denoting $y_{t} = 1 - \text{sg}(\bar{\mathcal{L}}_{\text{unique}}(\mathbf{z}_t; \theta))$ and $r_{t} = H_{\hat{\theta}}(\text{sg}(\mathbf{z}_{t}))$, where ``sg'' stands for stop gradients, we define the loss for training the model as
\begin{equation} \label{eq11}
        \mathcal{L}_{\text{filter}}(\mathbf{z}_{t}; \hat{\theta}) = -y_{t}\log r_{t}
        + (1-y_{t})\log(1 - r_{t}).
\end{equation}

\subsection{The Full Loss and Importance Scores}

With all the components, the loss for each frame in a video is:
\begin{equation} \label{eq12}
    \begin{aligned}
        \mathcal{L}(\mathbf{z}_{t}; \theta,\hat{\theta}) = \mathcal{L}_{\text{align}}(\mathbf{z}_{t}; \theta) + \lambda_1 \mathcal{L}_{\text{uniform}}(\mathbf{z}_{t}; \theta) \\
        + \lambda_2 \mathcal{L}_{\text{unique}}(\mathbf{z}_{t}; \theta) + \lambda_3 \mathcal{L}_{\text{filter}}(\mathbf{z}_{t}; \hat{\theta}),
    \end{aligned}
\end{equation}
where we fix both $\lambda_2$ and $\lambda_3$ as $0.1$ and only tune $\lambda_1$.

Scaling the local dissimilarity, global consistency, and uniqueness scores to $[0, 1]$ over $t$, the frame-level importance score is simply defined as:
\begin{equation}\label{eq13}
    p_{t} = \bar{\mathcal{L}}_{\text{align}}(\mathbf{z}_{t}; \theta)  \bar{\mathcal{L}}_{\text{uniform}}(\mathbf{z}_{t}; \theta)  \bar{H}_{\hat{\theta}}(\mathbf{z}_{t}) + \epsilon,
\end{equation}
which means that the importance scores will be high only when all three terms have nontrivial magnitude. $\epsilon$ is for avoiding zero values in the importance scores, which helps stabilize the knapsack algorithm for generating final summaries. As the scores are combined from three independent metrics, they tend not to have the temporal smoothness as possessed by the scores given by RNN \cite{zhang2016video} or attention networks \cite{fajtl2018summarizing}. We thus simply Gaussian-smooth the scores in each video to align with previous work in terms of temporal smoothness of scores.

%% file: experiment.tex
\section{Experiments}

\subsection{Datasets and Settings}\label{section:4.1}

\noindent
\textbf{Datasets.} Following previous work, we evaluate our method on two benchmarks: TVSum \cite{song2015tvsum} and SumMe \cite{gygli2014creating}. TVSum contains 50 YouTube videos, each annotated by 20 annotators in the form of importance scores for every two-second-long shot. SumMe includes 25 videos, each with 15-18 reference binary summaries. We follow \cite{zhang2016video} to use OVP (50 videos) and YouTube (39 videos) \cite{de2011vsumm} to augment TVSum and SumMe. Moreover, to test if our unsupervised approach can leverage a larger amount of videos, we randomly selected about 10,000 videos from the Youtube8M dataset \cite{abu2016youtube}, which has 3,862 video classes with highly diverse contents.

\noindent
\textbf{Evaluation Setting.} 
Again following previous work, we evaluate model performance using five-fold cross-validation, where the dataset (TVSum or SumMe) is randomly divided into five splits. The reported results are averaged over five splits. In the canonical setting (C) \cite{zhang2016video}, the training is only done on the original splits of the two evaluation datasets. In the augmented setting (A) \cite{zhang2016video}, we augment the training set in each fold with three other datasets (\eg, SumMe, YouTube, and OVP when TVSum is used for evaluation). In the transfer setting (T) \cite{zhang2016video}, all the videos in TVSum (or SumMe) are used for testing, and the other three datasets are used for training. Moreover, we introduce an extra transfer setting where the training is solely conducted on the collected Youtube8M videos, and the evaluation is done on TVSum or SumMe. This setting is designed for testing if our model can benefit from a larger amount of data.

\subsection{Evaluation Metrics}
\noindent
\textbf{F1 score}. Denoting $A$ as the set of frames in a ground-truth summary and $B$ as the set of frames in the corresponding generated summary, we can calculate the precision and recall as follows:
\begin{equation}\label{eq.10}
\text{Precision}=\frac{|A\cap B|}{|A|},\ \text{Recall}=\frac{|A\cap B|}{|B|},
\end{equation}
with which we can calculate the F1 score by
\begin{equation}\label{eq.11}
    \text{F1} = \frac{2\times \text{Precision}\times \text{Recall}}{\text{Precision} + \text{Recall}}.
\end{equation}
We follow \cite{zhang2016video} to deal with multiple ground-truth summaries and to convert importance scores into summaries. 

\noindent
\textbf{Rank correlation coefficients.} Recently, Otani \etal~\cite{otani2019rethinking} demonstrated that the F1 scores are unreliable and can be pretty high for even randomly generated summaries. They proposed to use rank correlation coefficients, namely Kendall's $\tau$ \cite{kendall1945treatment} and Spearman's $\rho$ \cite{beyer1991standard}, to measure the correlation between the predicted and the ground-truth importance scores. For each video, we first calculate the coefficient value between the predicted importance scores and each annotator's scores and then take the average over the total number of annotators for that video. The final results are obtained by averaging over all the videos. 

\subsection{Implementation Details} \label{section:4.3}
We follow previous work to use GoogleNet \cite{szegedy2015going} pre-trained features for standard experiments. For experiments with Youtube8M videos, we use the quantized Inception-V3 \cite{szegedy2016rethinking} features provided by the dataset \cite{abu2016youtube}. Both kinds of features are pre-trained on ImageNet \cite{krizhevsky2012imagenet}. The contrastive refinement module appended to the feature backbone is a lightweight Transformer encoder \cite{vaswani2017attention}, and so is the uniqueness filter. More architecture and training details can be found in Section \textcolor{red}{1} in the supplementary.

Following \cite{rochan2018video}, we restricted each video to have an equal length with random sub-sampling for longer videos and nearest-neighbor interpolation for shorter videos. Similar to \cite{rochan2018video}, we did not observe much difference when using different lengths, and we fixed the length to 200 frames, which is quite efficient for the training.

We tune two hyperparameters: The ratio $a$ that determines the size of the nearest neighbor set $\mathcal{N}_{t}$ and the coefficient $\lambda_1$ that controls the balance between the alignment and uniformity losses. Their respective effect will be demonstrated through the ablation study in Section \textcolor{red}{3} in the supplementary.


\subsection{Quantitative Results}
In this section, we compare our results with previous work and do the ablation study for different components of our method.

\textbf{Importance scores calculated with only pre-trained features} As shown in Tables~\ref{table1} and~\ref{table2}, $\bar{\mathcal{L}}^{*}_{\text{align}}$ and $\bar{\mathcal{L}}^{*}_{\text{uniform}}$ directly computed with GoogleNet \cite{szegedy2015going} pre-trained features already surpass most of the methods in terms of $\tau$, $\rho$ and F1 score. Especially, the correlation coefficient $\tau$ and $\rho$ surpass even supervised methods, \eg, (0.1345, 0.1776) v.s. dppLSTM's (0.0298, 0.0385) and SumGraph's (0.094, 0.138) for TVSum. Though $\text{DR-DSN}_{2000}$ has slightly better performance in terms of $\tau$ and $\rho$ for TVSum, it has to reach the performance after bootstrapping the online-generated summaries for $2000$ epochs while our results are directly obtained with simple computations using the same pre-trained features as those also used by DR-DSN.

\textbf{More training videos are needed for the contrastive refinement.} For the results in Tables~\ref{table1} and~\ref{table2}, the maximum number of training videos is only 159, coming from the SumMe augmented setting. For the canonical setting, the training set size is 40 videos for TVSum and 20 for SumMe. Without experiencing many videos, the model tends to overfit each specific video and cannot generalize well. This is similar to the observation in contrastive representation learning that a larger amount of data (from a larger dataset or obtained from data augmentation) helps the model generalize \cite{chen2020simple, caron2021emerging}. Therefore, the contrastive refinement results in Tables~\ref{table1} and~\ref{table2} hardly outperform those computed using pre-trained features.

\textbf{Contrastive refinement with Youtube8M on TVSum}. The model may better generalize to the test videos given sufficient training videos. This can be validated by the results for TVSum in Table~\ref{table3}. After the contrastive refinement, the results with only $\bar{\mathcal{L}}^{*}_{\text{align}}$ are improved from (0.0595, 0.0779) to (0.0911, 0.1196) for $\tau$ and $\rho$. We can also observe improvement over $\bar{\mathcal{L}}^{*}_{\text{align}} \& \bar{\mathcal{L}}^{*}_{\text{uniform}}$ brought by contrastive refinement. 

\begin{table}[h!]
\caption{Ablation results in terms of $\tau$ and $\rho$ together with their comparisons with previous work in the canonical setting. Since no previous work provided $\tau$ and $\rho$ for the other two settings, we provide our results for them in Section \textcolor{red}{2} in the supplementary. $\text{DR-DSN}_{60}$ means the DR-DSN trained for 60 epochs and similarly for $\text{DR-DSN}_{2000}$. Our scores with superscript $*$ are directly computed from pre-trained features. The results were generated with $(\lambda_1, a)=(0.5, 0.1)$. \textbf{Boldfaced} scores represent the best among supervised methods and human evaluations, and \textcolor{blue}{blue} scores are the best among the unsupervised methods. Please refer to the text for analyses of the results.}
\resizebox{\columnwidth}{!}{
\begin{tabular}{clcccc}
\toprule
& & \multicolumn{2}{c}{TVSum} & \multicolumn{2}{c}{SumMe} \\
\cmidrule(lr){3-4} \cmidrule(lr){5-6} 
& & $\tau$ & $\rho$ & $\tau$ & $\rho$ \\
\midrule
& Human baseline \cite{saquil2021multiple}                   & 0.1755           & 0.2019           & \textbf{0.1796}  & \textbf{0.1863} \\
\midrule
\multirow{4}{*}{ Supervised }
& VASNet \cite{fajtl2018summarizing, saquil2021multiple}     & 0.1690           & 0.2221           & 0.0224           & 0.0255  \\
& dppLSTM \cite{zhang2016video, otani2019rethinking}         & 0.0298           & 0.0385           & -0.0256          & -0.0311 \\
& SumGraph \cite{park2020sumgraph}                           & 0.094            & 0.138            & -                & -       \\
& Multi-ranker \cite{saquil2021multiple}                     & \textbf{0.1758}  & \textbf{0.2301}  & 0.0108           & 0.0137  \\
\midrule
\multirow{5}{*}{ Unsupervised (previous) } 
& $\text{DR-DSN}_{60}$ \cite{zhou2018deep, otani2019rethinking}            & 0.0169           & 0.0227           & 0.0433           & 0.0501  \\
& $\text{DR-DSN}_{2000}$ \cite{zhou2018deep, saquil2021multiple}      & 0.1516           & 0.198            & -0.0159          & -0.0218 \\
& $\text{SUM-FCN}_{unsup}$ \cite{rochan2018video, saquil2021multiple} & 0.0107           & 0.0142           &0.0080            & 0.0096  \\
& SUM-GAN \cite{mahasseni2017unsupervised, saquil2021multiple}                  
                                                             & -0.0535          & -0.0701          & -0.0095          & -0.0122 \\
& CSNet+GL+RPE \cite{jung2020global}                         & 0.070            & 0.091            & -                & -       \\
\midrule
\multirow{2}{*}{ Unsupervised (ours, w.o. training)} 
& $\bar{\mathcal{L}}^{*}_{\text{align}}$                        & 0.1055    & 0.1389  & \textcolor{blue}{0.0960}  & \textcolor{blue}{0.1173} \\

& $\bar{\mathcal{L}}^{*}_{\text{align}}$ $\&$ $\bar{\mathcal{L}}^{*}_{\text{uniform}}$     
                                                             & 0.1345           & 0.1776           & 0.0819           & 0.1001 \\
\midrule
\multirow{4}{*}{Unsupervised (ours, w. training)} 
& $\bar{\mathcal{L}}_{\text{align}}$                                            
                                                             & 0.1002           & 0.1321           & 0.0942           & 0.1151 \\
& $\bar{\mathcal{L}}_{\text{align}}$ $\&$ $\bar{\mathcal{L}}_\text{uniform}$                 
                                                             & 0.1231           & 0.1625           & 0.0689           & 0.0842 \\
& $\bar{\mathcal{L}}_{\text{align}}$ $\&$ $\bar{H}_{\hat{\theta}}$                 
                                                             & 0.1388           & 0.1827           & 0.0585           & 0.0715 \\
& $\bar{\mathcal{L}}_{\text{align}}$ $\&$ $\bar{\mathcal{L}}_\text{uniform}$ $\&$ $\bar{H}_{\hat{\theta}}$  
                              & \textcolor{blue}{0.1609}      & \textcolor{blue}{0.2118}           & 0.0358           & 0.0437 \\
\bottomrule
\end{tabular}
}
\label{table1}
\end{table}

\textbf{Contrastive refinement with Youtube8M on SumMe}. The reference summaries in SumMe are binary scores, and summary lengths are constrained to be within $15\%$ of the video lengths. Therefore, the majority part of the reference summary has exactly zeros scores. The contrastive refinement may still increase the scores' confidence for these regions, for which the annotators give zero scores thanks to the $15\%$ constraint. This eventually decreases the average correlation with the reference summaries, as per Table~\ref{table3}. 

However, suppose the predicted scores are refined to have sufficiently high confidence for regions with nonzero reference scores; in this case, they tend to be captured by the knapsack algorithm for computing the F1 scores. Therefore, we consider scores with both high F1 and high correlations to have high quality, as the former tends to neglect the overall correlations between the predicted and the annotated scores \cite{otani2019rethinking}, and the latter focuses on their overall ranked correlations but cares less about the prediction confidence.
This analysis may explain why the contrastive refinement for $\bar{\mathcal{L}}^{*}_{\text{align}}$ improves the F1 score but decreases the correlations. We analyze the negative effect of $\bar{\mathcal{L}}_{\text{uniform}}$ for SumMe later.

\begin{table}[h!]
\caption{Ablation results in terms of F1 together with their comparisons with previous unsupervised methods. The \textbf{boldfaced} results are the best ones. Please refer to Table \ref{table1}'s caption for the explanation of the notations and the text for analyses of the results.}
\resizebox{\columnwidth}{!}{
\begin{tabular}{lcccccc}
\toprule
& \multicolumn{3}{c}{TVSum} & \multicolumn{3}{c}{SumMe} \\
\cmidrule(lr){2-4} \cmidrule(lr){5-7}
& C & A & T & C & A & T \\
\midrule
$\text{DR-DSN}_{60}$ \cite{zhou2018deep}             & 57.6          & 58.4          & 57.8          & 41.4          & 42.8          & 42.4 \\
$\text{SUM-FCN}_{unsup}$ \cite{rochan2018video}      & 52.7          & -             & -             & 41.5          & -             & 39.5 \\
SUM-GAN \cite{mahasseni2017unsupervised}    & 51.7          & 59.5          & -             & 39.1          & 43.4          & - \\
UnpairedVSN \cite{rochan2019video}          & 55.6          & -             & 55.7          & 47.5          & -             & 41.6 \\
CSNet \cite{jung2019discriminative}         & 58.8          & 59            & 59.2          & \textbf{51.3} & \textbf{52.1} & 45.1 \\
CSNet+GL+RPE \cite{jung2020global}          & 59.1          & -             & -             & 50.2          & -             & - \\
$\text{SumGraph}_{unsup}$ \cite{park2020sumgraph}    & 59.3          & \textbf{61.2} & 57.6          & 49.8          & \textbf{52.1} & \textbf{47} \\
\midrule
$\bar{\mathcal{L}}^{*}_{\text{align}}$ 
                                            & 56.4          & 56.4          & 54.6          & 43.5          & 43.5          & 39.4 \\
$\bar{\mathcal{L}}^{*}_{\text{align}}$ $\&$ $\bar{\mathcal{L}}^{*}_{\text{uniform}}$    
                                            & 58.4          & 58.4          & 56.8          & 47.2          & 46.07         & 41.7 \\
\midrule
$\bar{\mathcal{L}}_{\text{align}}$ 
                                            & 54.6          & 55.1          & 53            & 46.8          & 47.1          & 41.5 \\
$\bar{\mathcal{L}}_{\text{align}}$ $\&$ $\bar{\mathcal{L}}_\text{uniform}$ 
                                            & 58.8          & 59.9          & 57.4          & 46.7          & 48.4          & 41.1 \\
$\bar{\mathcal{L}}_{\text{align}}$ $\&$ $\bar{H}_{\hat{\theta}}$  
                                            & 53.8          & 56            & 54.3          & 45.2          & 45            & 45.3 \\
$\bar{\mathcal{L}}_{\text{align}}$ $\&$ $\bar{\mathcal{L}}_\text{uniform}$ $\&$ $\bar{H}_{\hat{\theta}}$  
                                            & \textbf{59.5} & 59.9          & \textbf{59.7} & 46.8          & 45.5          & 43.9 \\
\bottomrule
\end{tabular}
}
\label{table2}
\end{table}

\textbf{The effect of $\bar{\mathcal{L}}_{\text{align}}$.} As can be observed in Tables \ref{table1},~\ref{table2}, and~\ref{table3}, solely using $\bar{\mathcal{L}}_{\text{align}}$ can already well quantify the frame importance. This indicates that $\bar{\mathcal{L}}_{\text{align}}$ successfully selects frames with diverse semantic information, which are indeed essential for a desirable summary. Moreover, we assume that diverse frames are a basis for a decent summary, thus always using $\bar{\mathcal{L}}_{\text{align}}$ for further ablations.

\textbf{The effect of $\bar{\mathcal{L}}_{\text{uniform}}$.} $\bar{\mathcal{L}}_{\text{uniform}}$ measures how consistent a frame is with the context of the whole video, thus helping remove frames with diverse contents but hardly related to the video theme. It is clearly demonstrated in Tables~\ref{table1} and \ref{table3} that incorporating $\bar{\mathcal{L}}_{\text{uniform}}$ helps improve the quality of the frame importance for TVSum. We thoroughly discuss why $\bar{\mathcal{L}}_{\text{uniform}}$ hurts SumMe performance in Section \textcolor{red}{7} of our supplementary.

\textbf{The effect of the uniqueness filter $\bar{H}_{\hat{\theta}}$.}  As shown in Tables~\ref{table1} and~\ref{table2}, though $\bar{H}_{\hat{\theta}}$ works well for TVsum videos, it hardly brings any benefits for the SumMe videos. Thus the good performance of the uniqueness filter for TVSum may simply stem from the fact that the background frames in TVSum are not challenging enough and can be easily detected by the uniqueness filter trained using only a few videos. Therefore, we suppose that $\bar{H}_{\hat{\theta}}$ needs to be trained on more videos in order to filter out more challenging background frames such that it can generalize to a wider range of videos. This is validated by the $\bar{\mathcal{L}}_{\text{align}}$\&$\bar{H}_{\hat{\theta}}$ results in Table~\ref{table3}, which indicate both decent F1 scores and correlation coefficients for both TvSum and SumMe. The TVSum performance can be further boosted when $\bar{\mathcal{L}}_{\text{uniform}}$ is incorporated. 

\begin{table}
\caption{The transfer evaluation setting with the Youtube8M dataset, where the training is solely conducted on the collected Youtube8M videos and then evaluated on TVSum and SumMe. The results from DR-DSN \cite{zhou2018deep} is also provided for comparison.}
\resizebox{\columnwidth}{!}{
\begin{tabular}{lcccccc}
\toprule
& \multicolumn{3}{c}{TVSum} & \multicolumn{3}{c}{SumMe} \\
\cmidrule(lr){2-4}\cmidrule(lr){5-7} 
& F & $\tau$ & $\rho$ & F & $\tau$ & $\rho$ \\
\midrule
DR-DSN \cite{zhou2018deep} & 51.6 & 0.0594 & 0.0788  & 39.8 & -0.0142   & -0.0176 \\
\midrule
$\bar{\mathcal{L}}^{*}_{\text{align}}$ & 55.9 & 0.0595 & 0.0779 & 45.5 & 0.1000 & 0.1237 \\
$\bar{\mathcal{L}}^{*}_{\text{align}}$ \& $\bar{\mathcal{L}}^{*}_\text{uniform}$  & 56.7 & 0.0680 & 0.0899 & 42.9 & 0.0531 & 0.0649 \\
\midrule
$\bar{\mathcal{L}}_{\text{align}}$ &  56.2 & 0.0911 & 0.1196 & 46.6 & 0.0776 & 0.0960 \\
$\bar{\mathcal{L}}_{\text{align}}$ \& $\bar{\mathcal{L}}_\text{uniform}$  & 57.3 & 0.1130 & 0.1490 & 40.9 & 0.0153 & 0.0190 \\
$\bar{\mathcal{L}}_{\text{align}}$ \& $\bar{H}_{\hat{\theta}}$ & 58.1 & 0.1230 & 0.1612 & 48.7 & 0.0780 & 0.0964 \\
$\bar{\mathcal{L}}_{\text{align}}$ \& $\bar{\mathcal{L}}_\text{uniform}$ \& $\bar{H}_{\hat{\theta}}$ & 59.4 & 0.1563 & 0.2048 & 43.2 & 0.0449 & 0.0553\\
\bottomrule
\end{tabular}
}
\label{table3}
\end{table}

\textbf{Comparison with DR-DSN \cite{zhou2018deep} on Youtube8M.} As per Table~\ref{table1}, DR-DSN is the only unsupervised method that has competitive performance with ours in terms of $\tau$ and $\rho$ and that has released the official implementation. We thus also trained DR-DSN on our collected Youtube8M videos to compare it against our method. As shown in Table~\ref{table3}, DR-DSN has a hard time generalizing to the evaluation videos. We also compare DR-DSN to our method with varying model sizes in Section \textcolor{red}{4} in the supplementary, where we compare the two methods' training efficiency and provide more training details as well.

\textbf{More experiments.} 
We perform the hyperparameter tuning on the challenging Youtube8M videos. We also evaluated the proposed metrics using different kinds of pre-trained features. Moreover, we observed that the F1 score for TVSum can be volatile to the importance scores' magnitude. The above results are in Sections \textcolor{red}{3}-\textcolor{red}{6} in the supplementary.

\begin{figure}
\begin{center}
\includegraphics[width=1.\linewidth]{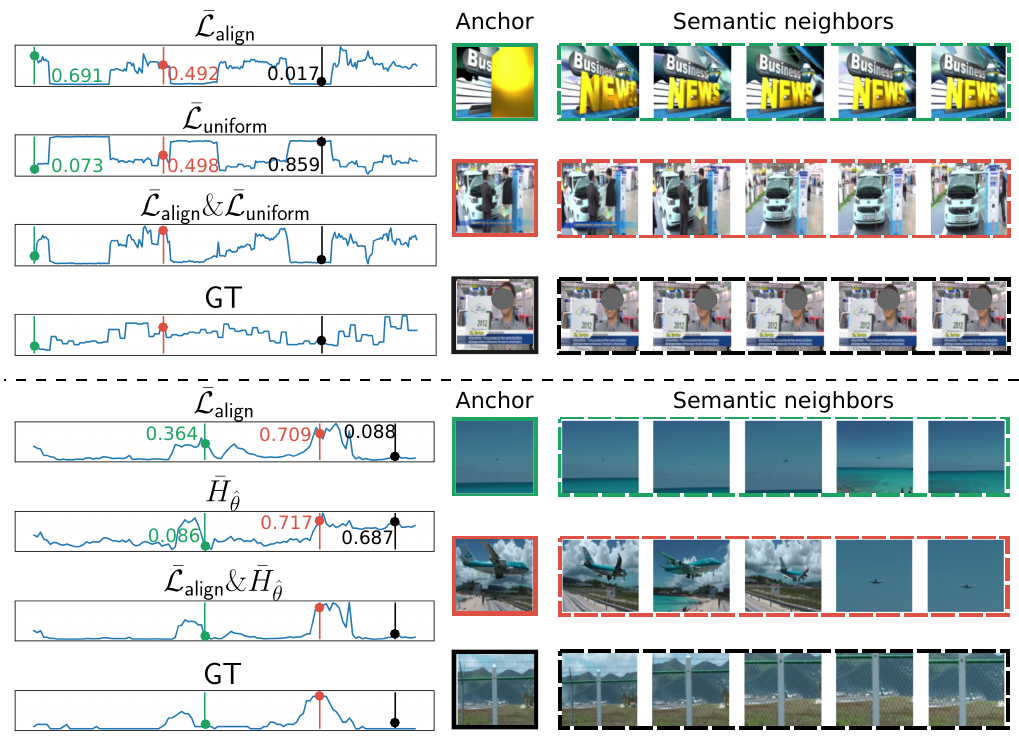}
\end{center}
\caption{The qualitative analysis for two video examples from TVSum and SumMe. The left column contains importance scores, where ``GT'' stands for ground truth. The \textcolor{mygreen}{green} bar selects an anchor frame with high $\bar{\mathcal{L}}_{\text{align}}$ but low $\bar{\mathcal{L}}_{\text{uniform}}$ or $\bar{H}_{\hat{\theta}}$, the \textcolor{myred}{red} bar selects one with nontrial magnitude for both metrics, and the \textbf{black} bar selects one with low $\bar{\mathcal{L}}_{\text{align}}$ but high $\bar{\mathcal{L}}_{\text{uniform}}$ or $\bar{H}_{\hat{\theta}}$. We show five samples from the top 10 semantic nearest neighbors within the dashed boxes on the right for each selected anchor frame.}
\label{fig3}
\end{figure}

\subsection{Qualitative Results}
We show the effect of the local dissimilarity ($\bar{\mathcal{L}}_{\text{align}}$), the global consistency ($\bar{\mathcal{L}}_{\text{uniform}}$), and the uniqueness scores generated by the uniqueness filter $\bar{H}_{\hat{\theta}}$ in Fig.~\ref{fig3}. We visualize and discuss the effects in pairs, \ie, $\bar{\mathcal{L}}_{\text{align}}$\&$\bar{\mathcal{L}}_{\text{uniform}}$ and $\bar{\mathcal{L}}_{\text{align}}$\&$\bar{H}_{\hat{\theta}}$. We provide an example of how $\bar{H}_{\hat{\theta}}$ improves $\bar{\mathcal{L}}_{\text{align}}$\&$\bar{\mathcal{L}}_{\text{uniform}}$ in Section \textcolor{red}{8} in the supplementary. In the upper half of Fig.~\ref{fig3}, the \textcolor{mygreen}{green} bar selects a frame with high local similarity but low global consistency, which turns out to be a title frame with a disparate appearance and hardly conveys any valuable information about the video. While the \textbf{black} bar selects a frame related to the main content of the video (an interview), it has semantic neighbors with almost the same look and is less likely to contain diverse semantics. The \textcolor{myred}{red} bar selects a frame with moderate local dissimilarity and global consistency. The frame, together with its semantic neighbors, conveys diverse information \eg, the car with or without people surrounding it. Moreover, it is also highly related to the whole video context: an interview in a car company. 

For the lower half of Fig.~\ref{fig3}, the \textcolor{mygreen}{green} bar selects a frame with information noticeably different from its neighbors, \eg, the sea occupies different proportions of the scene. However, such a frame can appear in any video with water scenes, rendering it not unique to the belonging video. Hence, its uniqueness score is low. The \textbf{black} bar selects a frame with an object specifically belonging to this video in the center, but the local semantic neighborhood around it hardly conveys very diverse information. The \textcolor{myred}{red} bar selects a frame with both high local dissimilarity and high uniqueness, which turns out to be the frame related to the gist of the video: St Maarten landing.

%% file: conclusion.tex
\section{Conclusion}
We take the first attempt to directly quantify frame-level importance without bootstrapping online-generated summaries for unsupervised video summarization based on the contrastive losses proposed by \cite{wang2020understanding}. With pre-trained deep features, our proposed metrics already yield high-quality importance scores compared to many previous heavily-trained methods. We can further improve the metrics with contrastive refinement by leveraging a sufficient amount of random videos. We also propose a novel uniqueness filter and validate its effectiveness through extensive experiments. It would be interesting to keep exploring the combination of unsupervised video summarization and representation learning. 

%% file: acknowledgement.tex
\section{Acknowledgement}
This work was partly supported by JST CREST Grant No.~JPMJCR20D3 and JST FOREST Grant No.~JPMJFR216O.

%% file: supp.tex
\onecolumn
\begin{center}
\textbf{\large Contrastive Losses Are Natural Criteria for Unsupervised Video Summarization: Supplementary Material}
\setcounter{equation}{0}
\setcounter{figure}{0}
\setcounter{table}{0}
\setcounter{page}{1}
\setcounter{section}{0}
\setcounter{enumiv}{0}

\end{center}
\section{Training Details} \label{section1}
We have used two kinds of pre-trained features in our experiments, namely the GoogleNet \citeSupp{szegedy2015goingSupp} features for the video summarization datasets and the quantized Inception-v3 \citeSupp{szegedy2016rethinkingSupp} features for the Youtube8M dataset, both with $1024$ dimensions. The GoogleNet features are provided by \citeSupp{zhang2016videoSupp} and the quantized Inception-v3 features by \citeSupp{abu2016youtubeSupp}. 

The model appended to the feature backbone for contrastive refinement is a stack of Transformer encoders with multi-head attention modules \citeSupp{vaswani2017attentionSupp}. There are two scenarios for our training: 1. The training with TVSum \citeSupp{song2015tvsumSupp}, SumMe \citeSupp{gygli2014creatingSupp}, YouTube, and OVP \citeSupp{de2011vsummSupp}, which is again divided into the canonical, augmented, and transfer settings; 2. The training with a subset of videos from Youtube8M dataset \citeSupp{abu2016youtubeSupp}. We call the training in the first scenario as \emph{Standard} and the second as \emph{YT8M}.

The pre-trained features are first projected into $128$ dimensions for training in both scenarios using a learnable, fully connected layer. The projected features are then fed into the Transformer encoders. The model architecture and associated optimization details are in Table~\ref{table1}.

The training for the $10,000$ Youtube8M videos takes around $6$ minutes for 40 epochs on a single NVIDIA RTX A6000. The efficient training benefits from the straightforward training objective (contrastive learning), lightweight model, the low dimensionality of the projected features, and the equal length of all the training videos (200 frames).

\begin{table}[h]
\caption{Model and optimization details used for the results in the main paper.}
\centering
\scalebox{0.8}{\begin{tabular}{lccccccccccc}
\toprule
& $\#$ Layers & $\#$ Heads & $\text{d}_{\text{model}}$ & $\text{d}_{\text{head}}$ & $\text{d}_{\text{inner}}$ & Optimizer & LR & Weight Decay & Batch Size & Epoch & Dropout \\
\midrule
Standard & 4 & 1 & 128 & 64 & 512 & Adam & 0.0001 & 0.0001 & 32 (TVSum) 8 (SumMe) & 40 & 0 \\
YT8M & 4 & 8 & 128 & 64 & 512 & Adam & 0.0001 & 0.0005 & 128 & 40 & 0 \\
\bottomrule
\end{tabular}}
\label{table1}
\end{table}

The ablation results for the numbers of encoder layers and attention heads are conducted on the Youtube8M videos alone, as they are much more challenging than the videos used in the standard training, and the results are in Section~\ref{section7}.

\section{The Full Evaluation Results for the Standard Training.}
In Table 1 in the main text, we only provided Kendall's $\tau$ and Spearman's $\rho$ for the canonical training setting, and we list the full results in Tables~\ref{table2} and ~\ref{table3}. For the analysis and the comparison with previous work, please refer to Table 1 in the main paper.

\begin{table}[h]
\caption{The full results for TVSum in the standard training.}
\centering
\scalebox{0.7}{
\begin{tabular}{clccccccccc}
\toprule
& & \multicolumn{3}{c}{Canonical} & \multicolumn{3}{c}{Augmented} & \multicolumn{3}{c}{Transfer} \\
\cmidrule(lr){3-5} \cmidrule(lr){6-8} \cmidrule(lr){9-11}
&& F1 & $\tau$ & $\rho$ & F1 & $\tau$ & $\rho$ & F1 & $\tau$ & $\rho$ \\
\midrule
\multirow{2}{*}{w.t. training} &  $\bar{\mathcal{L}}^{*}_{\text{align}}$ & 56.4 & 0.1055 & 0.1389  & 56.4 & 0.1055 & 0.1389 & 54.6  & 0.0956  & 0.1251 \\
& $\bar{\mathcal{L}}^{*}_{\text{align}} \& \bar{\mathcal{L}}^{*}_{\text{uniform}}$  & 58.4 & 0.1345 & 0.1776 & 58.4 & 0.1345 & 0.1776 & 56.8 & 0.1207 & 0.1589 \\
\midrule
\multirow{4}{*}{w. training} & $\bar{\mathcal{L}}_{\text{align}}$ & 54.6 & 0.1002 & 0.1321 & 55.1 & 0.1029 & 0.136 & 53.0 & 0.0831 & 0.1090 \\
& $\bar{\mathcal{L}}_{\text{align}} \& \bar{\mathcal{L}}_{\text{uniform}}$  & 58.8 & 0.1231 & 0.1625 & \textbf{59.9} &  0.1238 & 0.1631 & 57.4 & 0.1166 & 0.1529 \\
& $\bar{\mathcal{L}}_{\text{align}} \& \bar{H}_{\hat{\theta}}$ & 53.8 & 0.1388 & 0.1827 & 56 & 0.1363 & 0.1792 & 54.3 & 0.1173 & 0.1539 \\
& $\bar{\mathcal{L}}_{\text{align}} \& \bar{\mathcal{L}}_{\text{uniform}} \& \bar{H}_{\hat{\theta}}$ & \textbf{59.5} & \textbf{0.1609} & \textbf{0.2118} & 59.9 & \textbf{0.1623} & \textbf{0.2133} & \textbf{59.7} & \textbf{0.1405} & \textbf{0.1846} \\
\bottomrule
\end{tabular}}
\label{table2}
\end{table}

\begin{table}[h]
\caption{The full results for SumMe in the standard training.}
\centering
\scalebox{0.7}{
\begin{tabular}{clccccccccc}
\toprule
& & \multicolumn{3}{c}{Canonical} & \multicolumn{3}{c}{Augmented} & \multicolumn{3}{c}{Transfer} \\
\cmidrule(lr){3-5} \cmidrule(lr){6-8} \cmidrule(lr){9-11}
&& F1 & $\tau$ & $\rho$ & F1 & $\tau$ & $\rho$ & F1 & $\tau$ & $\rho$ \\
\midrule
\multirow{2}{*}{w.t. training} & $\bar{\mathcal{L}}^{*}_{\text{align}}$ & 43.5 & \textbf{0.0960} & \textbf{0.1173}& 43.5 & \textbf{0.0960} & \textbf{0.1173}  & 39.4 & \textbf{0.0769} & \textbf{0.0939}  \\
& $\bar{\mathcal{L}}^{*}_{\text{align}} \& \bar{\mathcal{L}}^{*}_{\text{uniform}}$  & 47.2  & 0.0819 & 0.1001 & 46.07  & 0.0819 & 0.1001 & 41.7 & 0.0597 & 0.073 \\
\midrule
\multirow{4}{*}{w. training} & $\bar{\mathcal{L}}_{\text{align}}$ & \textbf{46.8} & 0.0942 & 0.1151 & 47.1 & 0.0872 & 0.1065 & 41.5 & 0.0756 & 0.0924 \\
& $\bar{\mathcal{L}}_{\text{align}} \& \bar{\mathcal{L}}_{\text{uniform}}$  & 46.7 & 0.0689 & 0.0842 & \textbf{48.4} &  0.0645 & 0.0788 & 41.1 & 0.0305 & 0.0374 \\
& $\bar{\mathcal{L}}_{\text{align}} \& \bar{H}_{\hat{\theta}}$& 45.2 & 0.0585 & 0.0715 & 45 & 0.059 & 0.0721 & \textbf{45.3} & 0.0611 & 0.0747 \\
& $\bar{\mathcal{L}}_{\text{align}} \& \bar{\mathcal{L}}_{\text{uniform}} \& \bar{H}_{\hat{\theta}}$ & \textbf{46.8} & 0.0358 & 0.0437 & 45.5 & 0.0353 & 0.0431 & 43.9 & 0.0306 & 0.0374 \\
\bottomrule
\end{tabular}}
\label{table3}
\end{table}

\section{Hyperparameter Ablation Results.}

We focus on ablating two hyperparameters: $a$ for controlling the size of the nearest neighbor set $\mathcal{N}_{t}$ and $\lambda_1$ for balancing the alignment and uniformity losses. The ablation results are provided for when importance scores are produced by $\bar{\mathcal{L}}_{\text{align}} $\&$ \bar{H}_{\hat{\theta}}$ and by $\bar{\mathcal{L}}_{\text{align}} $\&$ \bar{H}_{\hat{\theta}} $\&$ \bar{\mathcal{L}}_{\text{uniform}}$.

\subsection{$\bar{\mathcal{L}}_{\text{align}}$ $\&$ $\bar{H}_{\hat{\theta}}$}
\begin{figure*}[ht!]
\centering
\includegraphics[width=0.7\linewidth]{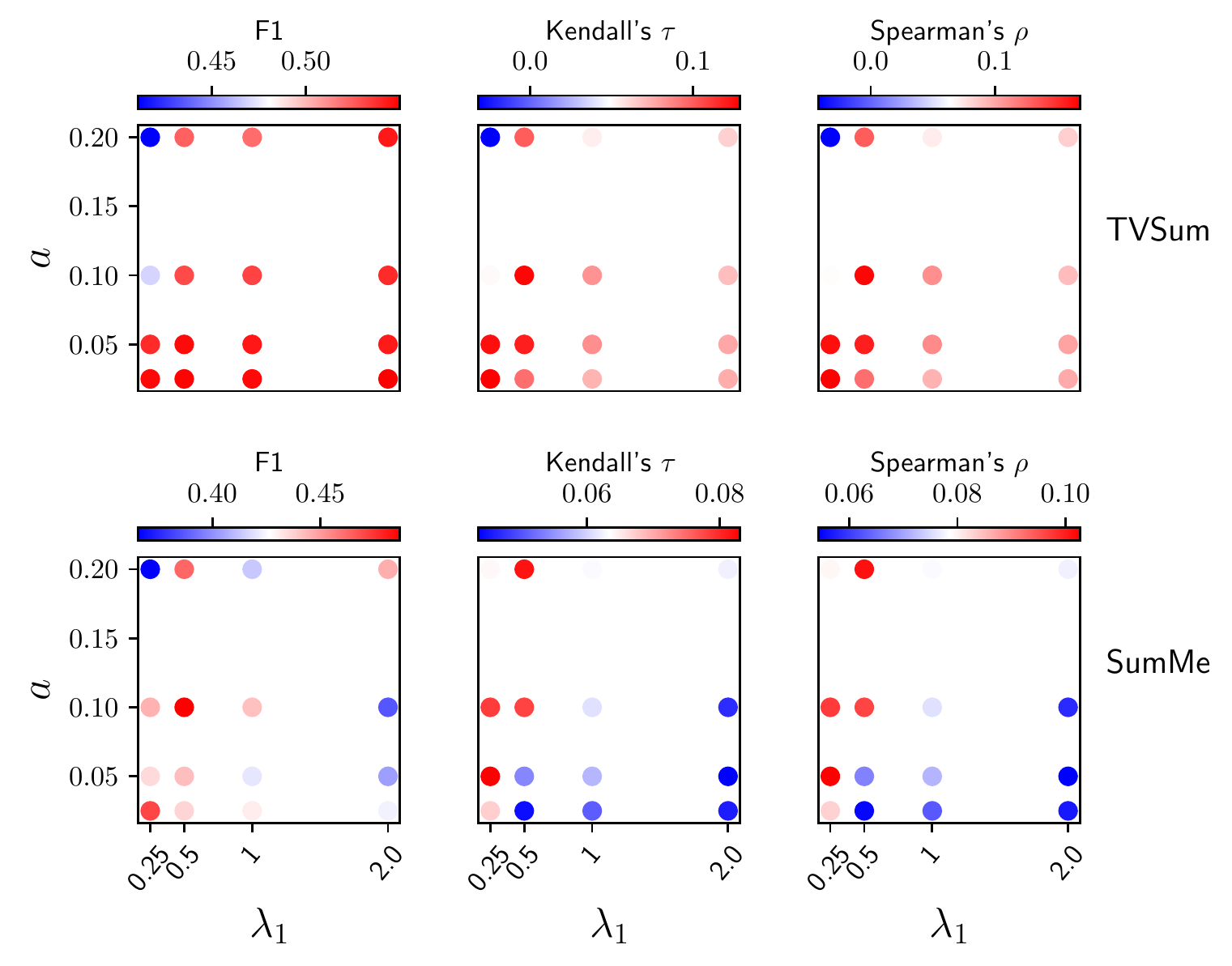}
\caption{Accompanying figure for Table~\ref{table5}}
\label{fig1}
\end{figure*}

As shown in Table~\ref{table5} and Fig.~\ref{fig1}, when $a$ becomes larger, TVSum performance begins to be unstable in terms of both F1 and correlation coefficients, and SumMe performance is relatively more stable, but also shows a similar unstable pattern in terms of F1. We hypothesize that when $a$ becomes larger, the nearest neighbor set becomes increasingly noisier, making both the alignment loss during training and the local dissimilarity metric (post-training alignment loss) for importance score generation less meaningful due to the semantically irrelevant neighbors.  For $\lambda_1$, smaller values generally give better performance when $a$ has a reasonable value, as larger values of $\lambda_1$ tend to make the uniformity loss suppress the alignment loss. Similarly, too small $\lambda_1$ will make the alignment loss suppress the uniformity loss, as we observed unstable training when further decreasing $\lambda_1$.

\begin{table}[ht!]
\caption{The ablation results for $a$ and $\lambda_1$ with $\bar{\mathcal{L}}_{\text{align}}$ $\&$ $\bar{H}_{\hat{\theta}}$ used for importance score calculation. See the text for analysis.} 
\centering
\scalebox{0.95}{
\begin{tabular}{cccccccc}
  \toprule
        \multicolumn{2}{c}{Hyper-params}       &  \multicolumn{3}{c}{TVSum}    &     \multicolumn{3}{c}{SumMe}  \\
            \cmidrule(lr){1-2}                       \cmidrule(lr){3-5}                  \cmidrule(lr){6-8} 
  $a$             & $\lambda_1$ &        F1 &     $\tau$ &   $\rho$ &        F1 &     $\tau$ &   $\rho$ \\
  \midrule
  \multirow{4}{*}{0.025} & 0.25 &      54.7 &     0.1286 &   0.1683 &      47.0 &     0.0671 &   0.0828 \\
                         & 0.50 &      54.9 &     0.0936 &   0.1227 &      43.7 &     0.0446 &   0.0550 \\
                         & 1.00 &      54.8 &     0.0717 &   0.0946 &      43.1 &     0.0508 &   0.0628 \\
                         & 2.00 &      55.0 &     0.0743 &   0.0981 &      42.3 &     0.0459 &   0.0567 \\
  \midrule                    
  \multirow{4}{*}{0.05}  & 0.25 &      53.8 &     0.1240 &   0.1624 &      43.5 &     0.0830 &   0.1027 \\
                         & 0.50 &      54.7 &     0.1189 &   0.1556 &      44.2 &     0.0540 &   0.0666 \\
                         & 1.00 &      54.4 &     0.0841 &   0.1110 &      42.0 &     0.0577 &   0.0714 \\
                         & 2.00 &      54.3 &     0.0764 &   0.1010 &      40.4 &     0.0436 &   0.0543 \\
  \midrule                       
  \multirow{4}{*}{0.1}   & 0.25 &      46.9 &     0.0498 &   0.0645 &      44.4 &     0.0783 &   0.0971 \\
                         & 0.50 &      52.9 &     0.1263 &   0.1655 &      48.7 &     0.0780 &   0.0964 \\
                         & 1.00 &      53.2 &     0.0829 &   0.1094 &      44.1 &     0.0609 &   0.0755 \\
                         & 2.00 &      53.8 &     0.0689 &   0.0907 &      38.6 &     0.0470 &   0.0583 \\
  \midrule                       
  \multirow{4}{*}{0.2}   & 0.25 &      41.1 &    -0.0318 &  -0.0421 &      36.6 &     0.0639 &   0.0793 \\
                         & 0.50 &      52.4 &     0.0994 &   0.1299 &      46.2 &     0.0816 &   0.1010 \\
                         & 1.00 &      52.0 &     0.0536 &   0.0707 &      41.3 &     0.0630 &   0.0781 \\
                         & 2.00 &      54.3 &     0.0631 &   0.0830 &      44.5 &     0.0622 &   0.0771 \\
\bottomrule
\end{tabular}}
\label{table5}
\end{table}

\subsection{$\bar{\mathcal{L}}_{\text{align}}$ $\&$ $\bar{H}_{\hat{\theta}}$ $\&$ $\bar{\mathcal{L}}_{\text{uniform}}$}
As shown in Table~\ref{table6}, the analysis is in general similar to that made in Table~\ref{table5}. However, we can observe that the performance has been obviously improved for TVSum but undermined for SumMe due to incorporating $\bar{\mathcal{L}}_{\text{uniform}}$. We will explain this phenomenon in Section~\ref{sec.luniform}. 

The results in the main paper for the YT8M training were produced for TVSum with $(a, \lambda_1)=(0.05, 0.25)$ and for SumMe with $(a, \lambda_1)=(0.1, 0.5)$, which are the best settings for the two datasets, respectively. This was done for avoiding the effect of the hyperparameters when we ablate the different components, \ie, $\bar{\mathcal{L}}_{\text{align}}$, $\bar{H}_{\hat{\theta}}$, and $\bar{\mathcal{L}}_{\text{uniform}}$. The results in the main paper for the standard training were produced with $(a, \lambda_1)=(0.1, 0.5)$, which is stable for both TVSum and SumMe when only few videos are available for training.

\begin{table}[ht]
\caption{The ablation results for $a$ and $\lambda_1$ with $\bar{\mathcal{L}}_{\text{align}}$ $\&$ $\bar{H}_{\hat{\theta}}$ $\&$ $\bar{\mathcal{L}}_{\text{uniform}}$ used for importance score calculation. See the text for analysis.} 
\centering
\scalebox{0.95}{
\begin{tabular}{cccccccc}
  \toprule
        \multicolumn{2}{c}{Hyper-params}       &  \multicolumn{3}{c}{TVSum}    &     \multicolumn{3}{c}{SumMe}  \\
            \cmidrule(lr){1-2}                       \cmidrule(lr){3-5}                  \cmidrule(lr){6-8} 
  $a$             & $\lambda_1$ &        F1 &     $\tau$ &   $\rho$ &        F1 &     $\tau$ &   $\rho$ \\
  \midrule
  \multirow{4}{*}{0.025} & 0.25 &      58.2 &     0.1403 &   0.1842 &      40.2 &     0.0341 &   0.0421 \\
                         & 0.50 &      57.3 &     0.0548 &   0.0716 &      38.7 &     0.0083 &   0.0101 \\
                         & 1.00 &      55.9 &    -0.0058 &  -0.0078 &      40.6 &    -0.0191 &  -0.0239 \\
                         & 2.00 &      54.5 &    -0.0087 &  -0.0112 &      40.3 &    -0.0069 &  -0.0087 \\
  \midrule                   
  \multirow{4}{*}{0.05}  & 0.25 &      58.5 &     0.1564 &   0.2050 &      42.7 &     0.0618 &   0.0765 \\
                         & 0.50 &      57.2 &     0.1205 &   0.1577 &      38.7 &     0.0182 &   0.0223 \\
                         & 1.00 &      55.5 &     0.0427 &   0.0563 &      39.7 &     0.0094 &   0.0114 \\
                         & 2.00 &      54.1 &     0.0074 &   0.0098 &      40.1 &    -0.0027 &  -0.0034 \\
  \midrule                       
  \multirow{4}{*}{0.1}   & 0.25 &      56.0 &     0.0743 &   0.0971 &      42.4 &     0.0737 &   0.0914 \\
                         & 0.50 &      57.8 &     0.1421 &   0.1866 &      43.2 &     0.0449 &   0.0553 \\
                         & 1.00 &      54.7 &     0.0446 &   0.0587 &      41.6 &     0.0130 &   0.0160 \\
                         & 2.00 &      54.7 &     0.0097 &   0.0127 &      41.3 &    -0.0142 &  -0.0176 \\
  \midrule                       
  \multirow{4}{*}{0.2}   & 0.25 &      50.9 &    -0.0267 &  -0.0355 &      41.8 &     0.0597 &   0.0740 \\
                         & 0.50 &      56.4 &     0.1178 &   0.1540 &      46.6 &     0.0626 &   0.0775 \\
                         & 1.00 &      50.7 &     0.0053 &   0.0067 &      39.0 &     0.0087 &   0.0109 \\
                         & 2.00 &      54.7 &     0.0162 &   0.0212 &      41.1 &    -0.0064 &  -0.0079 \\
\bottomrule

\end{tabular}}
\label{table6}
\end{table}

\begin{figure*}[hb!]
\centering
\includegraphics[width=0.7\linewidth]{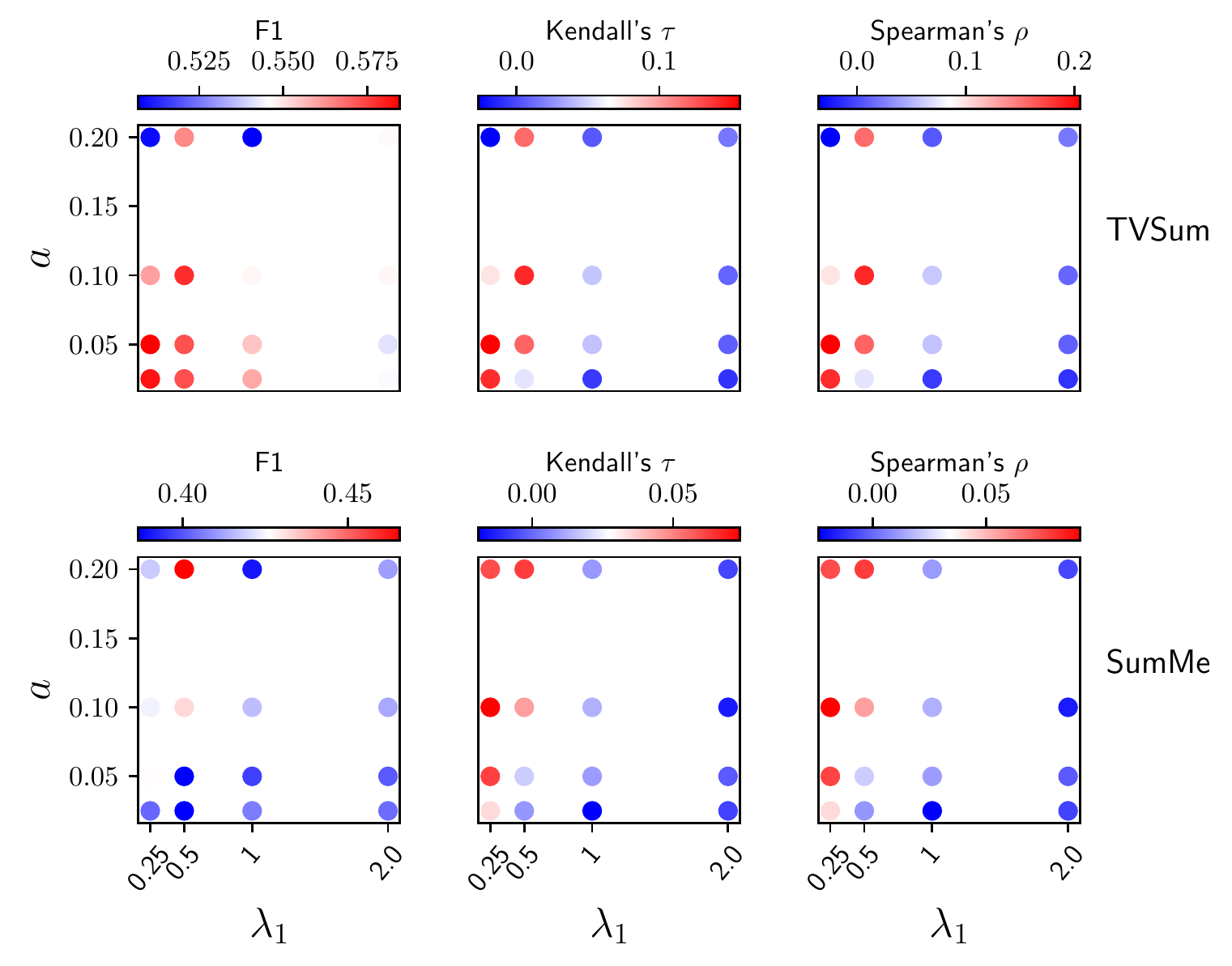}
\caption{Accompanying figure for Table~\ref{table6}}
\label{fig2}
\end{figure*}

\clearpage

\section{Ablation on Model Size and Comparison with DR-DSN on Youtube8M.}\label{section7}

In this section, we show the ablation results in Table~\ref{table7} for different sizes of the Transformer encoder \citeSupp{vaswani2017attentionSupp}, where the number of layers and the number of attention heads are varied. Meanwhile, we compare the results with those obtained from DR-DSN \citeSupp{zhou2018deepSupp} trained on the same collected Youtube8M videos, as DR-DSN has the best $\tau$ and $\rho$ among past unsupervised methods and is the only one that has a publicly available official implementation.

As can be observed in Table~\ref{table7}, the model performance is generally stable with respect to the model sizes, and we choose 4L8H as our default as reported in Section~\ref{section1}. Moreover, the DR-DSN has a hard time generalizing well to the test videos when trained on the Youtube8M videos.

We also recorded the training time on Youtube8M for our model and DR-DSN to show that our training objective based on the contrastive losses is much more efficient than DR-DSN's reinforcement learning-based one that bootstraps online-generated summaries. We chose DR-DSN for comparison in this regard because it is the most lightweight model (a single layer of bi-directional LSTM) and the only one that has officially released code among all the unsupervised methods. Our heaviest model (4L8H) only took around 10s for one pass through our Youtube8M dataset, and only took 40 epochs to complete the training. However, DR-DSN took 140s per epoch and 100 epochs to reach the performance in Table~\ref{table7}. The performance stayed stable if we kept training DR-DSN for more epochs. Both experiments were done on a single NVIDIA RTX A6000.

\begin{table}[ht]
\caption{Ablation results for the model size together with comparison with DR-DSN trained on the same Youtube8M videos, where 2L2H represents ``2 layers 2 heads'' and the rest goes similarly. All the three components $\bar{\mathcal{L}}_{\text{align}}$ $\&$ $\bar{H}_{\hat{\theta}}$ $\&$ $\bar{\mathcal{L}}_{\text{uniform}}$ are used with $(a, \lambda_1)=(0.05, 0.25)$ for both SumMe and TVSum for fair comparison with DR-DSN, which also uses a representativeness-based training objective.} 
\centering
\scalebox{1}{
\begin{tabular}{ccccccc}
\toprule
        &      \multicolumn{3}{c}{TVSum}    &     \multicolumn{3}{c}{SumMe} \\
                    \cmidrule(lr){2-4}                  \cmidrule(lr){5-7} 
        & F1 &     $\tau$ &   $\rho$ &        F1 &     $\tau$ &   $\rho$ \\
\midrule
DR-DSN \citeSupp{zhou2018deepSupp} &  51.62 &    0.0594 &   0.0788 &      39.82 &    -0.0142  & -0.0176 \\
\midrule
2L2H &                        58.0 &     0.1492 &   0.1953 &      42.9 &     0.0689 &   0.0850 \\
2L4H &                        58.1 &     0.1445 &   0.1894 &      42.8 &     0.0644 &   0.0794 \\
2L8H &                        58.8 &     0.1535 &   0.2011 &      44.0 &     0.0584 &   0.0722 \\
4L2H &                        57.4 &     0.1498 &   0.1963 &      45.3 &     0.0627 &   0.0776 \\
4L4H &                        58.3 &     0.1534 &   0.2009 &      43.1 &     0.0640 &   0.0790 \\
4L8H &                        58.5 &     0.1564 &   0.2050 &      42.7 &     0.0618 &   0.0765 \\
\bottomrule
\end{tabular}}
\label{table7}
\end{table}

\section{Comparing Different Pre-trained Features}
As our method can directly compute importance scores using pre-trained features, it is also essential for it to be able to work with different kinds of pre-trained features. To prove this, we computed and evaluated the importance scores generated with 2D supervised features, 3D supervised features and 2D self-supervised features in Table~\ref{table8}.

In general, different 2D features, whether supervised or self-supervised, all deliver decent results. Differences exist but are not huge. The conclusion made in the main text that $\bar{\mathcal{L}}_{\text{unif}}$ helps TVSum but undermines SumMe also holds for most of the features. Based on this, we conclude that as long as the features contain decent semantic information learned from either supervision or self-supervision, they are enough for efficient computation of the importance scores for video summarization. The performance of these features transferred to different downstream image tasks does not necessarily generalize to our method for video summarization, as the latter only requires reliable semantic information (quantified as dot products) to calculate heuristic metrics for video frames. After all, it may be reasonable to say that the linear separability of these features for image classification and transferability for object detection and semantic segmentation are hardly related to and thus unable to benefit a task as abstract as video summarization.

However, one interesting observation is that our method does not work well with 3D supervised video features. This is understandable since these 3D features were trained to encode the information of video-level labels, thus encoding less detailed semantic information in each frame on which our method is built. Still, such 3D features contain part of the holistic information of the associated video and may be a good vehicle for video summarization that can benefit from such information. We consider it interesting to incorporate 3D features into our approach and will explore it in our future work.

\begin{table}[ht]
\caption{Evaluation results with different pre-trained features. The results were produced under the transfer setting with $a=0.1$. } 
\centering
\scalebox{0.75}{
\begin{tabular}{llcccccccccccc}
\toprule
&& \multicolumn{6}{c}{TVSum} & \multicolumn{6}{c}{SumMe} \\
\cmidrule(lr){3-8}  \cmidrule(lr){9-14} 
& & \multicolumn{3}{c}{$\bar{\mathcal{L}}^{*}_{\text{align}}$} & \multicolumn{3}{c}{$\bar{\mathcal{L}}^{*}_{\text{align}}$ $\&$ $\bar{\mathcal{L}}^{*}_{\text{unif}}$} & \multicolumn{3}{c}{$\bar{\mathcal{L}}^{*}_{\text{align}}$} & \multicolumn{3}{c}{$\bar{\mathcal{L}}^{*}_{\text{align}}$ $\&$ $\bar{\mathcal{L}}^{*}_{\text{unif}}$} \\
\cmidrule(lr){3-5} \cmidrule(lr){6-8} \cmidrule(lr){9-11} \cmidrule(lr){12-14}
& & F1 & $\tau$ & $\rho$ & F1 & $\tau$ & $\rho$ & F1 & $\tau$ & $\rho$ & F1 & $\tau$ & $\rho$ \\
\midrule
\multirow{8}{*}{ Supervised (2D) } & VGG19 \citeSupp{simonyan2014verySupp}         &        50.62 &         0.0745 &         0.0971 &             55.91 &              0.1119 &              0.1473 &        45.16 &         0.0929 &         0.1151 &             43.28 &              0.0899 &              0.1114 \\
& GoogleNet \citeSupp{szegedy2015goingSupp}     &        54.67 &         0.0985 &         0.1285 &             57.09 &              0.1296 &              0.1699 &        41.89 &         0.0832 &         0.1031 &             40.97 &              0.0750 &              0.0929 \\
& InceptionV3 \citeSupp{szegedy2016rethinkingSupp}  &        55.02 &         0.1093 &         0.1434 &             55.63 &              0.0819 &              0.1082 &        42.71 &         0.0878 &         0.1087 &             42.30 &              0.0688 &              0.0851 \\
& ResNet50 \citeSupp{he2016deepSupp}         &        51.19 &         0.0806 &         0.1051 &             55.19 &              0.1073 &              0.1410 &        42.30 &         0.0868 &         0.1076 &             43.86 &              0.0737 &              0.0914 \\
& ResNet101 \citeSupp{he2016deepSupp}        &        51.75 &         0.0829 &         0.1081 &             54.88 &              0.1118 &              0.1469 &        42.32 &         0.0911 &         0.1130 &             44.39 &              0.0736 &              0.0913 \\
& ViT-S-16 \citeSupp{dosovitskiy2020imageSupp} &        53.48 &         0.0691 &         0.0903 &             56.15 &              0.1017 &              0.1332 &        40.30 &         0.0652 &         0.0808 &             40.88 &              0.0566 &              0.0701 \\
& ViT-B-16 \citeSupp{dosovitskiy2020imageSupp}   &        52.85 &         0.0670 &         0.0873 &             56.15 &              0.0876 &              0.1152 &        42.10 &         0.0694 &         0.0860 &             41.65 &              0.0582 &              0.0723 \\
& Swin-S \citeSupp{liu2021swinSupp}  &        52.05 &         0.0825 &         0.1082 &             57.58 &              0.1120 &              0.1475 &        41.18 &         0.0880 &         0.1090 &             41.63 &              0.0825 &              0.1022 \\
\midrule
\multirow{2}{*}{ Supervised (3D) } & R3D50 \citeSupp{hara3dcnnsSupp}         &        52.09 &         0.0590 &         0.0766 &             53.35 &              0.0667 &              0.0869 &        37.40 &         0.0107 &         0.0138 &             41.03 &              0.0150 &              0.0190 \\
& R3D101 \citeSupp{hara3dcnnsSupp}       &        49.77 &         0.0561 &         0.0727 &             52.15 &              0.0644 &              0.0834 &        33.62 &         0.0173 &         0.0216 &             34.96 &              0.0212 &              0.0264 \\
\midrule
\multirow{5}{*}{ Self-supervised (2D) } & MoCo \citeSupp{he2020momentumSupp}          &        51.31 &         0.0797 &         0.1034 &             55.97 &              0.1062 &              0.1390 &        42.01 &         0.0768 &         0.0953 &             43.19 &              0.0711 &              0.0882 \\
& DINO-S-16 \citeSupp{caron2021emergingSupp} &        52.50 &         0.0970 &         0.1268 &             57.57 &              0.1200 &              0.1583 &        42.77 &         0.0848 &         0.1050 &             42.67 &              0.0737 &              0.0913 \\
& DINO-B-16 \citeSupp{caron2021emergingSupp} &        52.48 &         0.0893 &         0.1170 &             57.02 &              0.1147 &              0.1515 &        41.07 &         0.0861 &         0.1066 &             44.14 &              0.0679 &              0.0843 \\
& BEiT-B-16 \citeSupp{bao2021beitSupp}  &        49.64 &         0.1125 &         0.1468 &             56.34 &              0.1270 &              0.1665 &        36.91 &         0.0554 &         0.0686 &             38.48 &              0.0507 &              0.0629 \\
& MAE-B-16 \citeSupp{he2022maskedSupp}   &        50.40 &         0.0686 &         0.0892 &             54.58 &              0.1013 &              0.1327 &        40.32 &         0.0560 &         0.0695 &             39.46 &              0.0484 &              0.0601 \\
\bottomrule
\end{tabular}}
\label{table8}
\end{table}

\section{Unstable F1 scores}
We observed that F1 values could be highly unstable for TVSum. When there are zeros in the predicted importance scores, the F1 values will be unreasonably low for TVSum but relatively stable for SumMe. This happens because when all segments have non-zero scores, the knapsack algorithm favors selecting short segments, which is how the ground-truth summaries for TVSum were created \citeSupp{zhang2016videoSupp}. With such ground-truth summaries, the predicted importance scores without zero values can already have a good starting point for the F1 value since they basically go through the same knapsack algorithm with the short-segment bias used to create the ground-truth. Similar observations were made in \citeSupp{otani2019rethinkingSupp}. Segments with zero scores will not be selected by the knapsack algorithm and make the problem harder to solve, thus leading to low F1 values for TVSum. The SumMe summaries are not created from the knapsack and directly come from the users. Therefore, the F1 values on SumMe are less affected.

To make a fair comparison with previous work, which usually does not have zeros in the predicted scores, we chose to add a small value $\epsilon$ (\eg, 0.05) to the predicted scores. However, this does not address the unstableness of the F1 values.

Let us consider another choice to avoid zero scores: $\exp(p - 1)$, where $p\in [0, 1]$ is the predicted importance score for a frame. We obtain the results in Table \ref{table9} by using these two choices without changing any other part of the method.

As can be observed in Table \ref{table9}, merely shifting or scaling the importance scores without changing their relative magnitude can lead to different F1 values for both TVSum and SumMe. At the same time, the correlation coefficients are much more stable. Thus, we reinforced the conclusion in \citeSupp{otani2019rethinkingSupp} that the F1 value is not as stable as the correlation coefficients.

\begin{table}[ht]
\caption{The results obtained with different strategies for avoiding zero values in the predicted importance scores. $\epsilon$ stands for the operation $p+\epsilon$ and $\exp$ for $\exp(p-1)$, where $p\in [0,1]$ is the importance score for a frame.} 
\centering
\scalebox{0.9}{
\begin{tabular}{lcccccccccccc}
\toprule
& \multicolumn{6}{c}{TVSum} & \multicolumn{6}{c}{SumMe} \\
\cmidrule(lr){2-7} \cmidrule(lr){8-13}
& \multicolumn{2}{c}{F} & \multicolumn{2}{c}{$\tau$} & \multicolumn{2}{c}{$\rho$} & \multicolumn{2}{c}{F1} & \multicolumn{2}{c}{$\tau$} & \multicolumn{2}{c}{$\rho$} \\
\cmidrule(lr){2-3}\cmidrule(lr){4-5}\cmidrule(lr){6-7}\cmidrule(lr){8-9}\cmidrule(lr){10-11}\cmidrule(lr){12-13}
 & $\exp$ & $\epsilon$ & $\exp$ & $\epsilon$ & $\exp$ & $\epsilon$ & $\exp$ & $\epsilon$ & $\exp$ & $\epsilon$ & $\exp$ & $\epsilon$\\
\midrule
$\bar{\mathcal{L}}_{\text{align}} \& \bar{H}_{\hat{\theta}}$ & 58.1 & 53.8 & 0.123 & 0.124 & 0.1612 & 0.1624 & 46 & 48.7 & 0.0776 & 0.078 & 0.0958 & 0.0964 \\
$\bar{\mathcal{L}}_{\text{align}} \& \bar{H}_{\hat{\theta}} \& \bar{\mathcal{L}}_{\text{uniform}}$ & 59.4 & 58.5 & 0.1563 & 0.1564 & 0.2048 & 0.205 & 42.86 & 43.2 & 0.0441 & 0.0449 & 0.0544 & 0.0553 \\
\bottomrule
\end{tabular}}
\label{table9}
\end{table}

\section{The Behavior of $\bar{\mathcal{L}}^{*}_{\text{uniform}}$ on SumMe.}\label{sec.luniform}
In the main text, we saw that $\bar{\mathcal{L}}^{*}_{\text{uniform}}$ improved performance for TVSum but hurt for SumMe. We now discuss why this happened.

Firstly, when frames with zero annotated scores (recall the 15$\%$ constraint for SumMe) have high $\bar{\mathcal{L}}^{*}_{\text{align}}$, $\bar{\mathcal{L}}^{*}_{\text{uniform}}$ will reinforce such false confidence as long as it is not low enough to almost zero out $\bar{\mathcal{L}}^{*}_{\text{align}}$. Such frames may not necessarily be highly irrelevant and have low $\bar{\mathcal{L}}^{*}_{\text{uniform}}$ to begin with because most of them, though still informative, are eliminated by the $15\%$ constraint. As shown in Table~\ref{table10}, keeping the most confidence values of $\bar{\mathcal{L}}^{*}_{\text{align}}$ by simple thresholding, which may help remove false predictions of $\bar{\mathcal{L}}^{*}_{\text{align}}$, can mitigate this issue. The results in the main text are all without such thresholding for a fair comparison with previous work.

\begin{figure*}[ht!]
\centering
\includegraphics[width=0.5\linewidth]{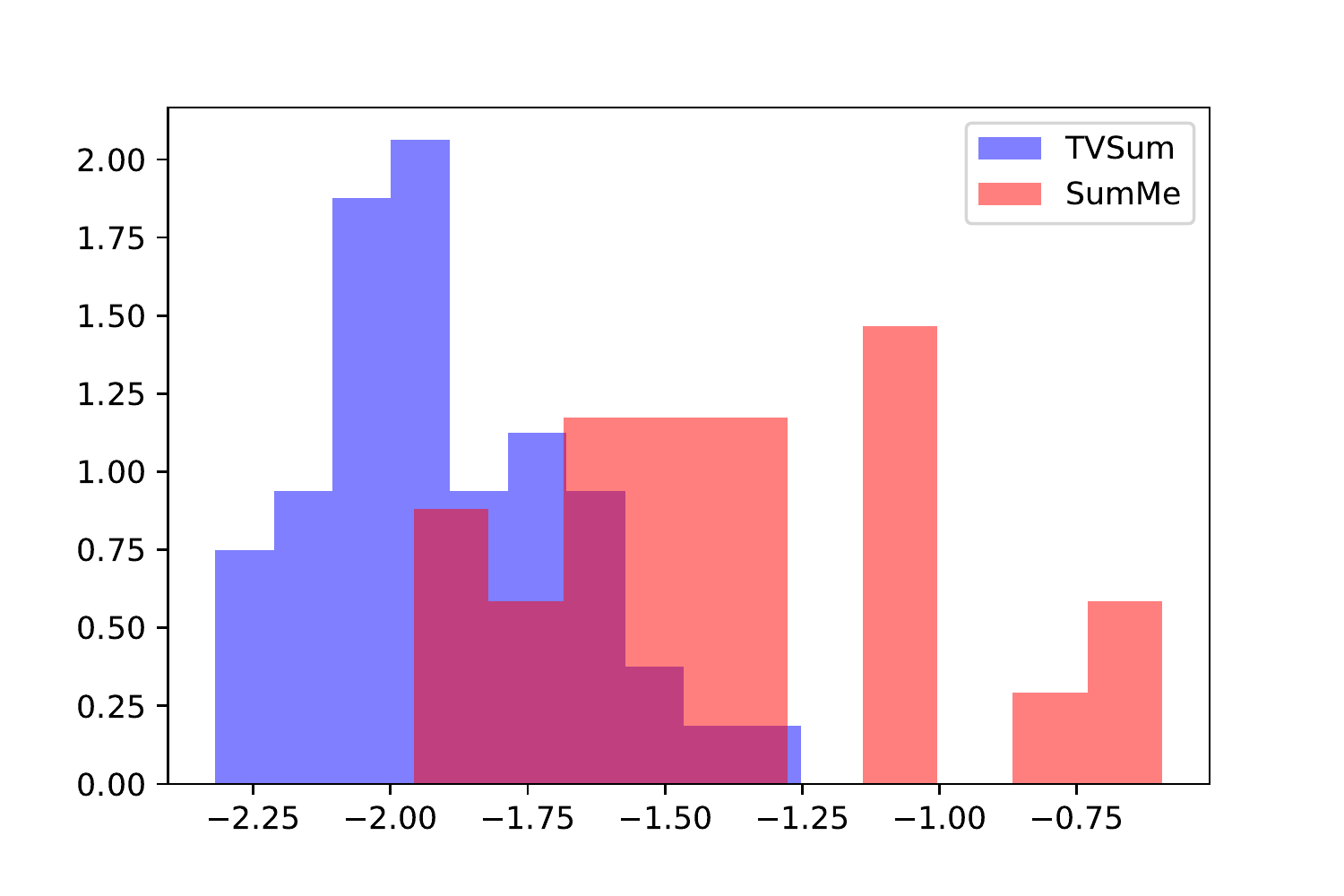}
\caption{The histogram (density) of $\bar{\mathcal{L}}^{*}_{\text{uniform}}$ (before normalization) for TVSum and SumMe videos. It is clear that SumMe videos have distinctly higher values than those for TVSum videos.}
\label{fig3}
\end{figure*}

\begin{figure}
     \centering
     \begin{subfigure}[b]{0.6\textwidth}
         \centering
         \includegraphics[width=\textwidth]{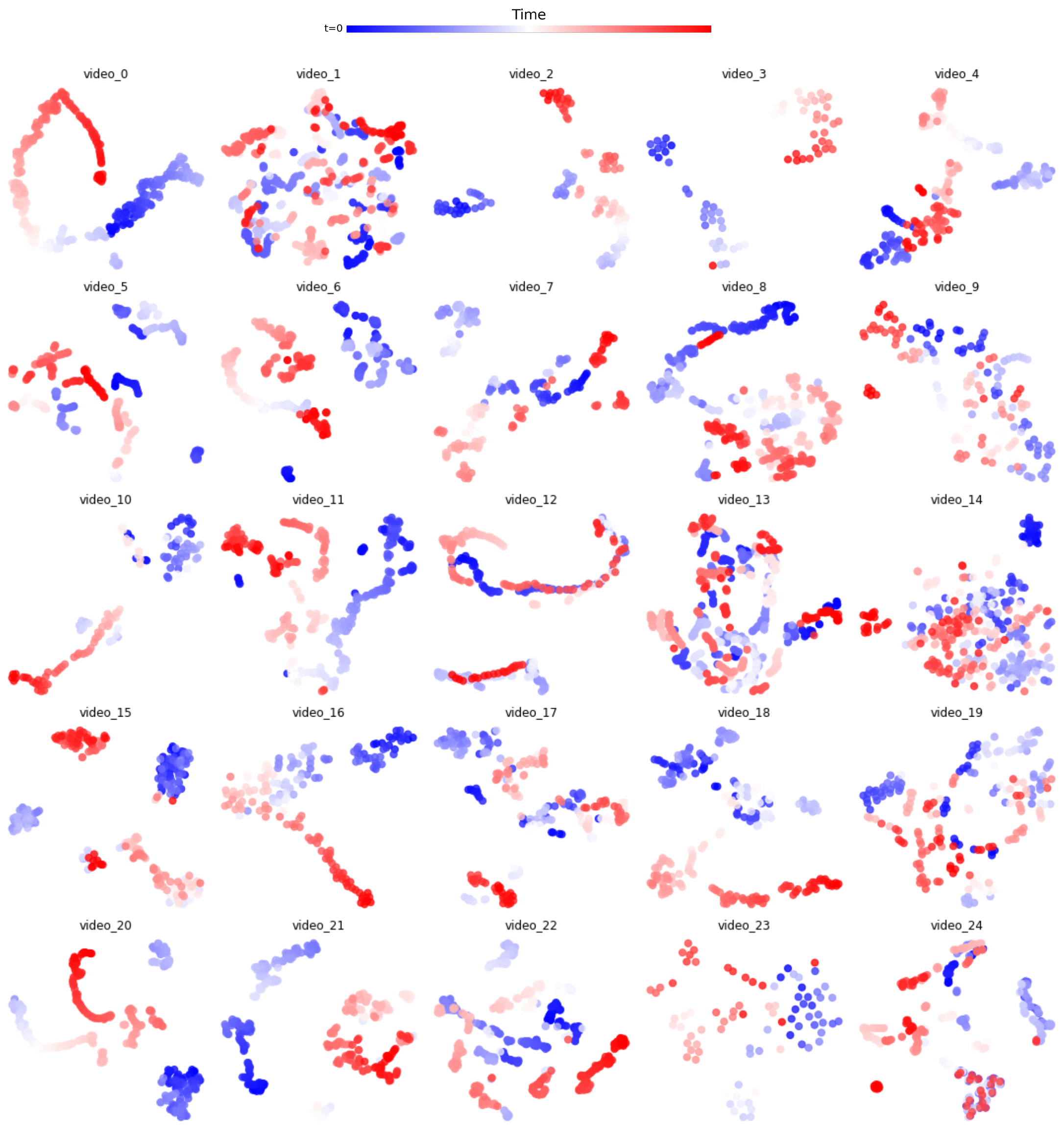}
         \caption{T-SNE plots for all the SumMe videos.}
         \label{fig4a}
     \end{subfigure}
     \hfill
     \begin{subfigure}[b]{0.6\textwidth}
         \centering
         \includegraphics[width=\textwidth]{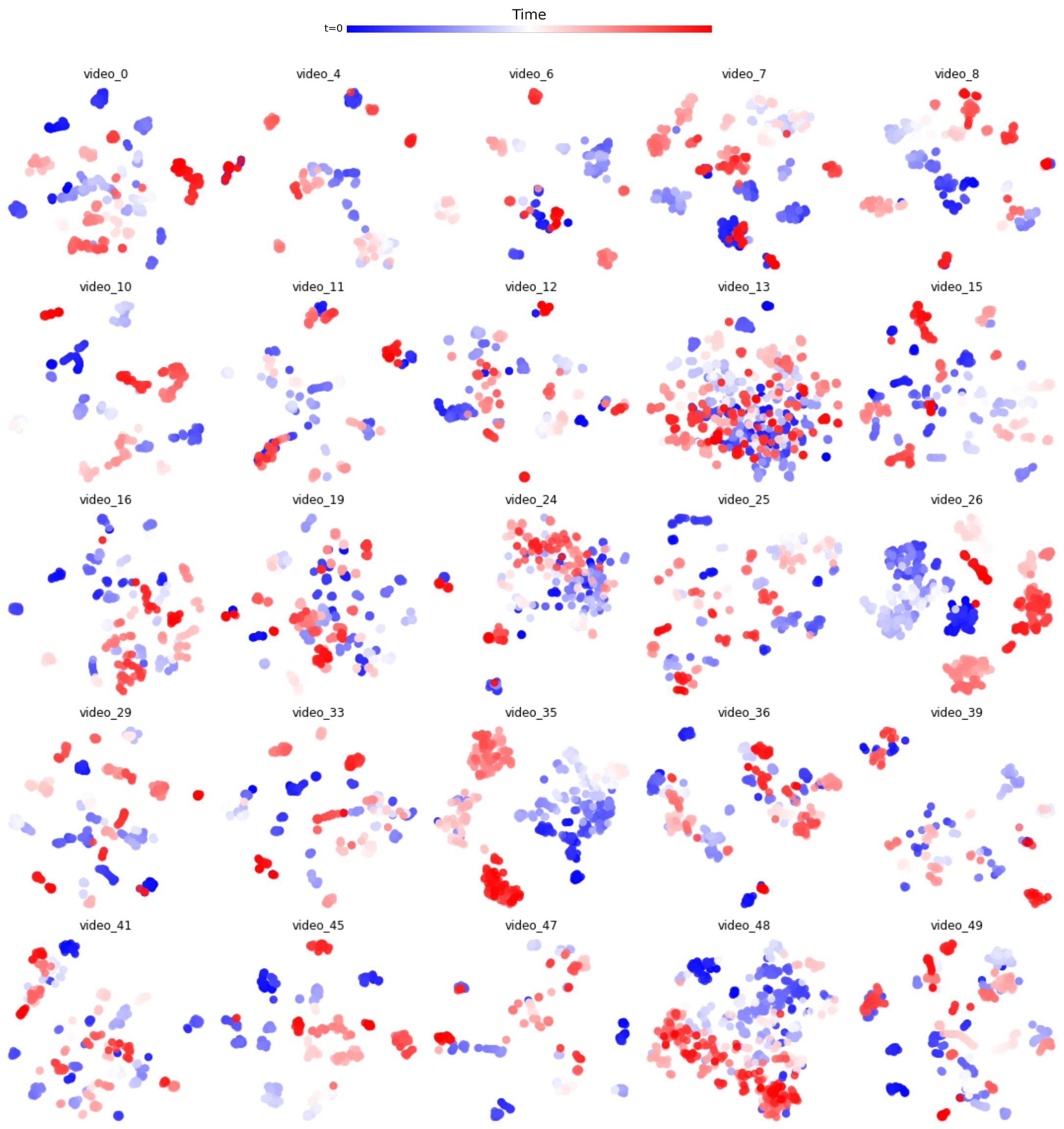}
         \caption{T-SNE plots for randomly selected 25 TVSum videos.}
         \label{fig4b}
     \end{subfigure}
     \caption{TSNE plots for SumMe and TVSum videos. See text for analysis.}
\label{fig4}
\end{figure}

Though the results in Table~\ref{table10} show improvement for $\bar{\mathcal{L}}^{*}_{\text{uniform}}$, it is not as striking as that for TVSum in terms of $\tau$ and $\rho$.
Compared to TVSum videos, many videos in SumMe already contain quite consistent frames due to their slowly evolving properties. Such slowly evolving features can be visualized by T-SNE plots shown in Fig.~\ref{fig4a} under the comparison against Fig.~\ref{fig4b}. For videos with such consistent contents, the $\mathcal{L}^{*}_{\text{uniform}}$ (before normalization) tend to be high for most of the frames. We show the normalized histogram of $\mathcal{L}^{*}_{\text{uniform}}$ for both TVSum and SumMe videos in Figure~\ref{fig3}. As can be observed, SumMe videos have distinctly higher $\mathcal{L}^{*}_{\text{uniform}}$ than those of TVSum videos. 

Consequently, for videos already possessing consistent contents, $\mathcal{L}^{*}_{\text{uniform}}$ normalized to 0 and 1 ($\bar{\mathcal{L}}^{*}_{\text{uniform}}$) will filter out frames that are relatively the least consistent. However, such frames are still relevant and may well likely possess a certain level of diversity, which the annotators of SumMe favor (see the results for $\bar{\mathcal{L}}^{*}_{\text{align}}$ for SumMe in Table 1, 2 and 3 in the main text). Thus, removing such frames using the $\bar{\mathcal{L}}^{*}_{\text{uniform}}$ will cause harm to the correlations. For some videos with truly noisy frames, $\bar{\mathcal{L}}^{*}_{\text{uniform}}$ retains its functionality of attenuating their importance, leading to better correlations. Thus, the pros and cons brought by $\bar{\mathcal{L}}^{*}_{\text{uniform}}$ may cancel out on average for the whole SumMe dataset, eventually yielding minor improvement in correlation (\ie $\tau$ and $\rho$). $\bar{\mathcal{L}}^{*}_{\text{uniform}}$ thus suits better complex videos such as those in TVSum, since such videos may inevitably contain noisy frames that can be well detected by $\bar{\mathcal{L}}^{*}_{\text{uniform}}$.

\begin{table}[ht]
\caption{The results were obtained from the transfer setting on SumMe directly with pre-trained features with $a=0.1$. Thresholding means simply setting $\bar{\mathcal{L}}^{*}_{\text{align}}[\bar{\mathcal{L}}^{*}_{\text{align}}<\alpha \text{AVG}(\bar{\mathcal{L}}^{*}_{\text{align}})]=0$, where AVG means taking average over the whole video and $\alpha=1.1$.}
\centering
\scalebox{1.}{
\begin{tabular}{ccccccc}
\toprule
& \multicolumn{3}{c}{Before thresholding} & \multicolumn{3}{c}{After thresholding} \\
\cmidrule(lr){2-4} \cmidrule(lr){5-7}
& F & $\tau$ & $\rho$ & F & $\tau$ & $\rho$ \\
\midrule
$\bar{\mathcal{L}}^{*}_{\text{align}}$ & 39.4 & 0.0769 & 0.0939 & 40.7 & 0.0969 & 0.1119 \\
$\bar{\mathcal{L}}^{*}_{\text{align}} \& \bar{\mathcal{L}}^{*}_{\text{uniform}}$  & 41.7 & 0.0597 & 0.073 & 44.5 & 0.0982 & 0.1134 \\
\bottomrule
\end{tabular}}
\label{table10}
\end{table}

\section{Complete Figures for the Qualitative Analysis}

\begin{figure*}[ht!]
\centering
\includegraphics[width=1\linewidth]{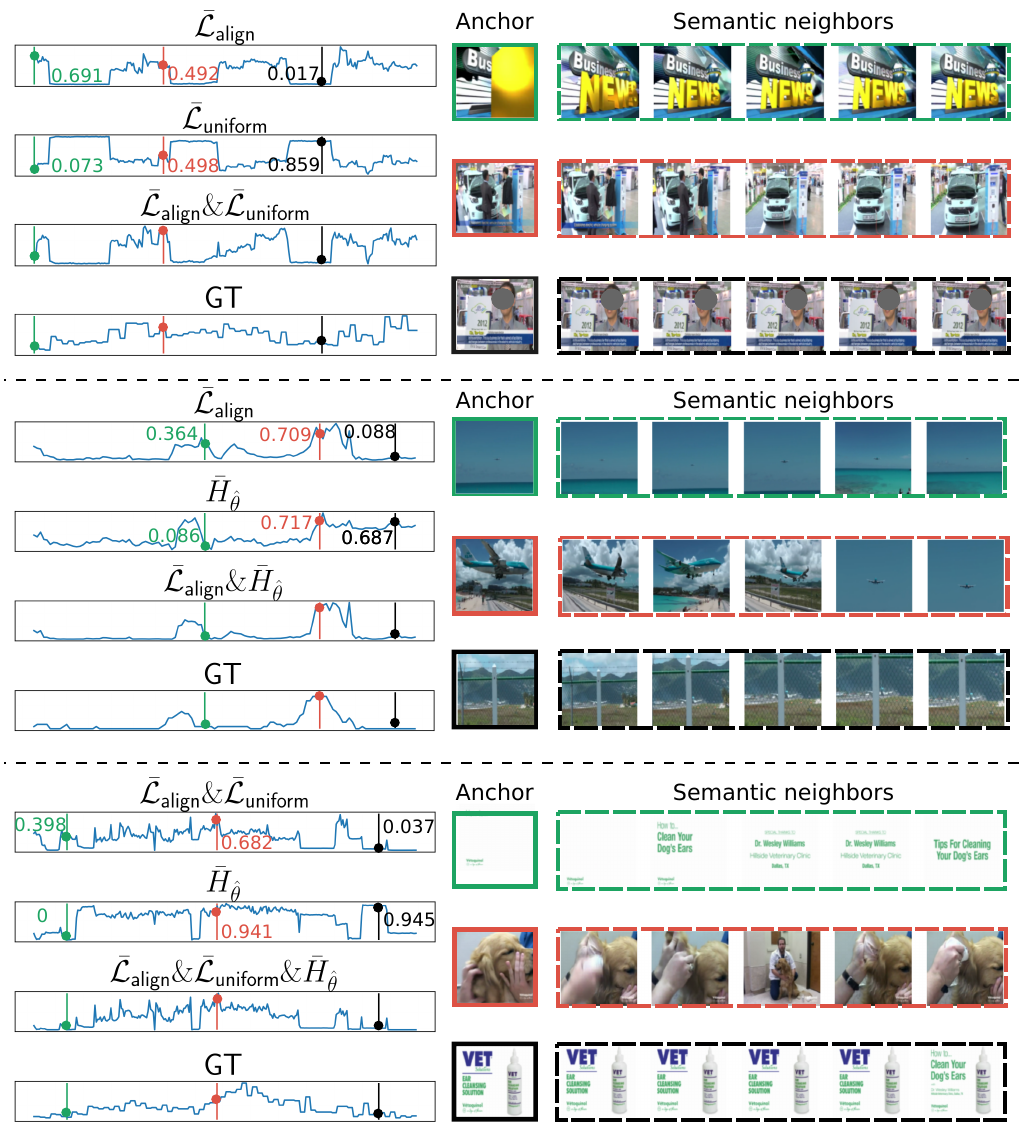}
\caption{The complete figure for the qualitative analysis in the main paper.All the important scores are scaled to $[0,1]$ for visualization.}
\label{fig5}
\end{figure*}

In Fig.\ref{fig5}, we provide a larger version of the figure for the qualitative analysis in the main paper (the top and the middle) together with another one showing how $\bar{H}_{\hat{\theta}}$ improves $\bar{\mathcal{L}}_{\text{align}} \& \bar{\mathcal{L}}_{\text{uniform}}$ (bottom). Please refer to the main paper for the analyses of the first two figures.

As shown in the bottom figure of Fig.~\ref{fig5}, the \textcolor{mygreen}{green} bar selects a frame with nontrivial local dissimilarity and global consistency but low uniqueness. It turns out that the frame contains only texts and has semantic neighbors with different texts, which can lead to high local dissimilarity. As the video is an instruction video, such frames with only textual illustrations should be prevalent, thus also incurring nontrivial global consistency. However, such textual frames are common in nearly all kinds of instruction videos or other videos with pure textual frames, hence the low uniqueness score. The \textcolor{myred}{red} bar selects a frame where some treatment is ongoing for a dog, which is specific to the current video and satisfies the local dissimilarity and the global consistency. The \textbf{black} bar selects a frame with low local dissimilarity and global consistency but high uniqueness, which is a frame showing off the vet medicine. Though quite specific to the video, the frame is not favored by the other two metrics.

